\pdfoutput=1
\documentclass[11pt]{article}
\usepackage[final]{acl}

\usepackage{times}
\usepackage{latexsym}
\usepackage{booktabs}
\usepackage[T1]{fontenc}
\usepackage[utf8]{inputenc}
\usepackage{microtype}
\usepackage{inconsolata}
\usepackage{graphicx}
\usepackage{marvosym}
\usepackage{hyperref}
\usepackage{algpseudocode}
\usepackage{xcolor,colortbl}
\usepackage{amssymb}
\usepackage{colortbl}
\definecolor{mygray}{gray}{.90}
\usepackage{multirow}
\usepackage{pifont}

\usepackage{amssymb}
\usepackage{mathtools}
\usepackage{amsthm}
\usepackage[capitalize,noabbrev]{cleveref}
\theoremstyle{plain}

\theoremstyle{definition}

\theoremstyle{remark}

\usepackage[textsize=tiny]{todonotes}

\title{MapNav: A Novel Memory Representation via Annotated Semantic Maps for Vision-and-Language Navigation}

\author{ Lingfeng Zhang$^{1,2,3,}$\thanks{~Co-first Authors. \\ \indent~{lfzhang715@gmail.com, xshao@baai.ac.cn} \\\indent~$^\dag$ Project Leader. \\\indent~\textsuperscript{\Letter} {Corresponding Authors.}  \\ \indent~{renjingxu@hkust-gz.edu.cn, shanghang@pku.edu.cn
}}~~~~~~~ {Xiaoshuai Hao$^{2,*,\dag}$}~~~~~~~ {Qinwen Xu$^{2,6}$} ~~~~~~~ {Qiang Zhang$^{1,4}$}~~~~~~~\\ {{\bf Xinyao Zhang$^{1}$} ~~~~~ {\bf Pengwei Wang$^{2}$}~~~~~  {\bf {Jing Zhang$^{5}$}}~~~~~  {\bf Zhongyuan Wang$^{2}$}~~~~~} \\ {{\bf Shanghang Zhang$^{2,6,}\textsuperscript{\Letter}$}~~~~~ {\bf Renjing Xu$^{1,}\textsuperscript{\Letter}$}}  \vspace{5pt} \\
{$^1$The Hong Kong University of Science and Technology (Guangzhou)} \\{$^2$ Beijing Academy of Artificial Intelligence, $^3$ Tsinghua University} \\ {$^4$Beijing Innovation Center of Humanoid Robotics Co., Ltd.} \\ {$^5$ School of Computer Science, Wuhan University} \\ {$^6$ State Key Laboratory of Multimedia Information Processing,} \\{ School of Computer Science, Peking University}} 

\begin{document}
\maketitle
\null\vspace{10pt}
\begin{abstract}

Vision-and-language navigation (VLN) is a key task in Embodied AI, requiring agents to navigate diverse and unseen environments while following natural language instructions. 
Traditional approaches rely heavily on historical observations as spatio-temporal contexts for decision making, leading to significant storage and computational overhead.
In this paper, we introduce \textbf{\textit{MapNav}}, a novel end-to-end VLN model that leverages Annotated Semantic Map (ASM) to replace historical frames.
Specifically, our approach constructs a top-down semantic map at the start of each episode and update it at each timestep, allowing for precise object mapping and structured navigation information. 
Then, we enhance this map with explicit textual labels for key regions, transforming abstract semantics into clear navigation cues and generate our ASM.
\textbf{\textit{MapNav}} agent using the constructed ASM as input, and use the powerful end-to-end capabilities of VLM to empower VLN.
Extensive experiments demonstrate that \textbf{\textit{MapNav}} achieves state-of-the-art (SOTA) performance in both simulated and real-world environments, validating the effectiveness of our method.
Moreover, we will release our ASM generation source code and dataset to ensure reproducibility, contributing valuable resources to the field.
We believe that our proposed \textbf{\textit{MapNav}} can be used as a new memory representation method in VLN, paving the way for future research in this field.
\end{abstract}

\section{Introduction}
\label{sec1:into}

\begin{figure}[!ht]
\centering
\null\vspace{25pt}
\includegraphics[width=0.46\textwidth]{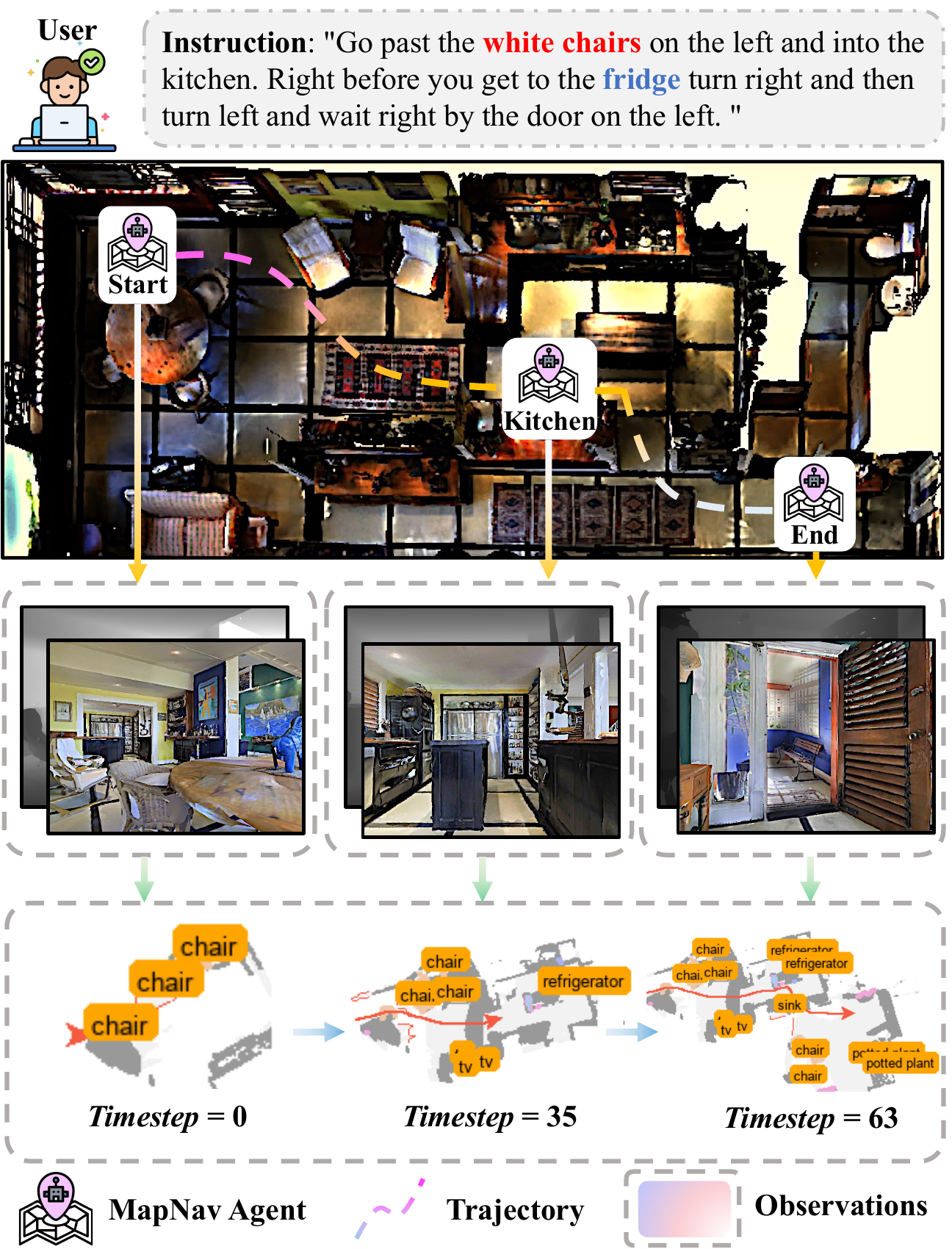}
\vspace{5pt}
\caption{
\textbf{Illustration of our Annotated Semantic
Map (ASM).} 
At each timestep, MapNav agent leverages egocentric observations to capture semantic objects and assign explicit textual labels to key regions, creating the ASM for the current moment.
ASM provides information such as physical obstacles, explored regions, the agent’s current position, trajectory and semantic objects.
}
\vspace{-10pt}
\label{fig1}
\end{figure}

Vision-and-language Navigation (VLN)~\cite{gu2022vision,park2023visual} is a crucial integration of embodied AI and multimodal understanding~\cite{HAO2025103018, ji2025robobrain, tang2025affordgrasp}, empowering autonomous agents to interpret natural language instructions and navigate complex, unseen environments.
Unlike traditional navigation tasks that rely solely on visual input or predetermined waypoints, 
VLN requires a complex blend of language understanding and visual perception.
For example, agents must interpret subtle instructions~\cite{vasudevan2021talk2nav, an2024etpnav, chen2024affordances} such as ``\textit{walk past the potted plants and turn left in front of the wooden cabinet}'', while also processing dynamic visual scenes and making real-time navigation decisions.

Existing VLN methods can be classified into discrete and continuous navigation paradigms, primarily distinguished by their representations of action space. Discrete environment navigation methods leverage MP3D~\cite{chang2017matterport3d} and abstract the navigation space into a connected graph structure for waypoint selection. While these methods have demonstrated impressive performance on benchmark datasets~\cite{shah2023lm, zhou2024navgpt,long2024discuss}, they fail to reflect the continuity of real-world navigation. To address this gap, the Habitat simulator~\cite{savva2019habitat} and benchmarks like R2R-CE~\cite{krantz2020beyond} and RxR-CE~\cite{ku2020room} have enabled research in continuous navigation. Moreover, to minimize the sim-to-real gap, Habitat implements a low-level action space with forward movement and rotational actions.
Existing continuous environment navigation methods~\cite{zhang2024navid, zheng2024towards} heavily rely on historical robot observations as spatio-temporal contexts for decision-making and instruction following. 
However, these approaches significantly increase storage requirements and lack a structured understanding of past trajectories.
Therefore, designing a novel memory representation to effectively replace traditional historical frames is of great significance and thus becomes the motivation for our work.

To fill this gap, we propose a novel end-to-end VLM-based VLN model, \textbf{\textit{MapNav}}, which leverages Annotated Semantic Maps for innovative memory representation, effectively replacing traditional historical frames.
Specifically, we first transform RGB-D and pose data into point cloud representations to generate precise top-down visualizations. 
Then, we align semantic segmentation to construct a base semantic map, which is further enhanced into an Annotated Semantic Map (ASM) by integrating explicit textual annotations for salient regions and abstract semantic concepts.
As shown in Fig.~\ref{fig1}, the ASM is initialized at the beginning of each episode and updated at each timestep for real-time operation. 
It encapsulates crucial environmental information for navigation, including physical obstacle distributions, explored areas, agent position, historical trajectories, and semantic object locations. 
The \textbf{\textit{MapNav}} agent utilizes the constructed ASM as input, harnessing the powerful end-to-end capabilities of VLM to enhance vision-and-language navigation.
Extensive experiments highlight the pivotal role of ASM in VLN tasks, showcasing memory representation capabilities that are comparable to traditional historical frame-based approaches. 
Additionally, ASM enables VLMs to develop structured spatial understanding and optimize path selection effectively. These advantages make ASM a promising method for enhancing navigation performance in various environments. We will release a dataset containing \textbf{1 Million} step-wise samples, featuring RGB frames, corresponding ASMs, VLN instructions, and frame-specific actions to promote this field. We believe that our proposed \textbf{\textit{MapNav}} can serve as an innovative memory representation method in VLN, paving the way for future advancements in this field.
Our main contributions are summarized as follows:
\vspace{-5pt}
\begin{itemize}

\item 
We propose a novel end-to-end VLM-based vision-and-language navigation model, \textbf{\textit{MapNav}}, which leverages Annotated Semantic Maps for innovative memory representation, effectively replacing traditional historical frames.
\vspace{-5pt}
\item 
We introduce a top-down Annotated Semantic Map (ASM) that we update at each timestep, enabling precise object mapping and structured navigation, while enhancing it with explicit textual labels for key regions to provide clear navigation cues.
\vspace{-5pt}
\item 
\textbf{\textit{MapNav}} outperforms state-of-the-art (SOTA) methods in both simulated and real-world environments, offering a new memory representation in VLN and paving the way for future research in the field.
\vspace{-5pt}
\item 
We will release our ASM generation source code and dataset, allowing for the reproducibility of the results presented in this study, which will serve as a valuable contribution to the field.

\vspace{-10pt}
\end{itemize}
\begin{figure*}[t]
\centering
\includegraphics[width=0.92\textwidth]{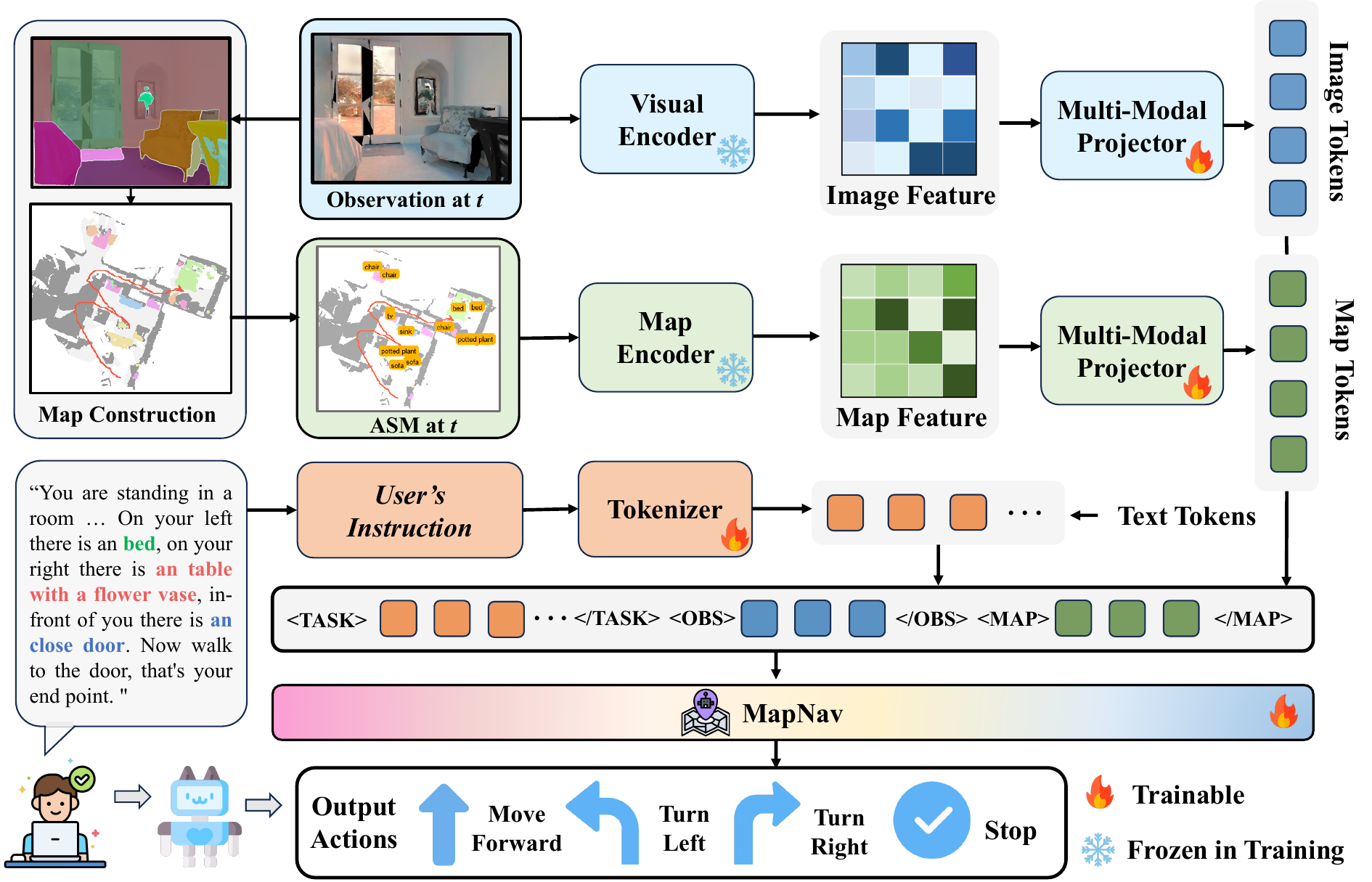}
\caption{\textbf{An overview of MapNav framework.} 
We present a top-down Annotated Semantic Map (ASM), updated at each timestep for precise object mapping and structured navigation. It features explicit textual labels for key regions, providing clear navigation cues. The current RGB observation, ASM, and instruction are used as inputs to an end-to-end VLM framework, which generates navigation actions in natural language.
}
\vspace{-10pt}
\label{fig2}
\end{figure*}

\section{Related Work}
\label{sec2:related}
\textbf{Vision-and-Language Navigation}  
Vision-and-Language Navigation (VLN) \cite{gu2022vision,park2023visual} in embodied AI~\cite{li2024foundation,zhang2025vtla,hao2025tla} focuses on navigating unseen environments by following human instructions, primarily in discretized simulated scenarios \cite{anderson2020rxr, thomason2020cvdn}. Agents navigate between predefined nodes on a graph by integrating language and visual input \cite{qi2020object, liu2024volumetric}, but this reliance complicates real-world deployment.
To address this, VLNs for Continuous Environments (VLN-CE) \cite{krantz2020beyond, savva2019habitat} enable unrestricted navigation through low-level control or waypoint-based methods \cite{hong2022bridging, krantz2021waypoint}, improving sim-to-real transferability despite added complexity.
Recent advancements in vision-language models have significantly influenced VLN development, utilizing large-scale pre-trained models \cite{zhang2024navid, zheng2024towards, wuevaluating} and VLN-specific pre-training \cite{hao2020prevalent, wu2022cross}. For instance, NavGPT \cite{zhou2024navgpt} autonomously generates actions using GPT-4o, while DiscussNav \cite{long2024discuss} employs VLN experts to reduce human involvement. InstructNav \cite{long2024instructnav} decomposes navigation into subtasks, and Nav-CoT \cite{lin2024navcot} uses Chain of Thoughts (CoT) \cite{wei2022chain} for environmental simulation. Some methods \cite{zheng2024towards, zhang2024navid} fine-tune VLMs for specific navigation tasks, highlighting flexibility. However, existing approaches often depend on hierarchical prompts or historical frames, leading to high memory demands and limited understanding of past data.
This paper introduces a novel memory representation using Annotated Semantic Maps (ASMs) to effectively replace traditional historical frames.

\noindent\textbf{Map Representations for VLN} 
Structured maps in VLN enhance navigation performance by improving environmental understanding \cite{wang2023gridmm, hong2023ego2map, hao2024mapdistill, hao2025msc,hao2024mbfusion}. 
Methods like MC-GPT \cite{zhan2024mc} and VoroNav \cite{wu2024voronav} utilize topological maps to capture viewpoints and spatial relationships, while InstructNav \cite{long2024instructnav} and VLFM \cite{yokoyama2024vlfm} create value maps for waypoint selection.
Semantic maps \cite{zhang2024trihelper, zhang2024multi, hong2023ego2map, yu2023l3mvn} retain object-level information for navigation. However, these maps are often not interpretable by VLMs.
To address this, we propose Annotated Semantic
Map (ASM), a novel semantic map representation that allows VLMs to explicitly understand rich map information, including obstacle distributions, explored areas, agent positions, historical trajectories, and semantic object locations. 
ASM aims to establish a new memory representation for VLN.

\begin{figure*}[t]
\centering
\includegraphics[width=0.96\textwidth]{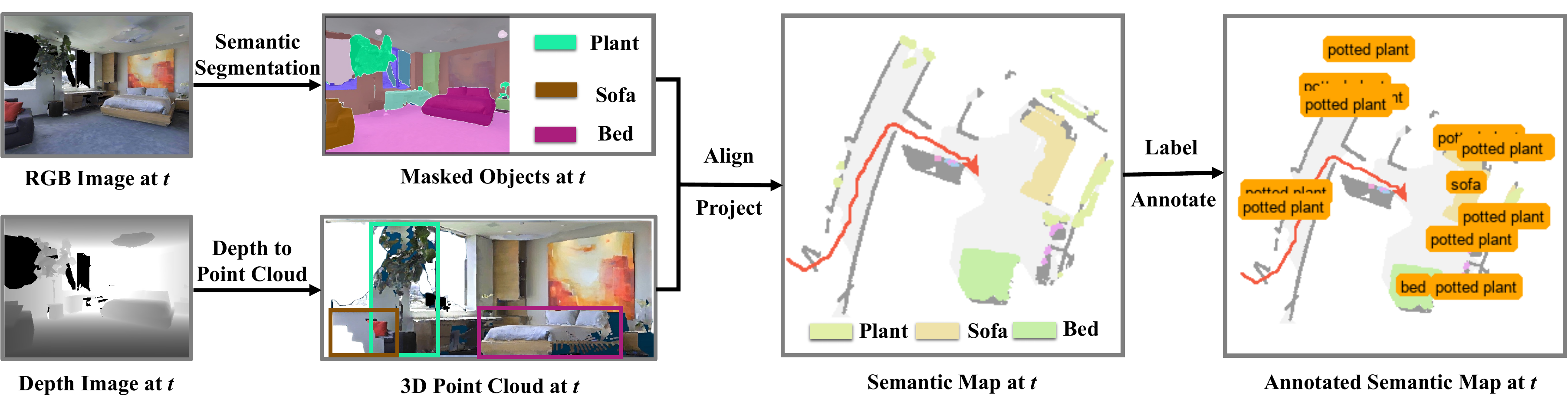}
\caption{\textbf{ASM Generation Process.}
Semantic map generation starts with episode initialization. 
At each timestep, the RGB image is processed by a semantic segmentation module to create a semantic mask aligned with the depth-converted 3D point cloud. By combining this with the previous pose transformation, we project the 3D point cloud onto a 2D plane to update the semantic map. Finally, we convert the semantic map into the ASM through region clustering and text annotation, yielding a comprehensive memory representation with labeled objects.
}
\vspace{-12pt}
\label{fig3}
\end{figure*}
\section{Methodology}
MapNav pursues a novel end-to-end vision-and-language navigation model based on VLM, leveraging Annotated Semantic Maps (ASM) to replace traditional historical frames. 
We introduce a top-down annotated semantic map at the start of each episode, updating it at each timestep for precise object mapping and structured navigation information, while enhancing it with textual labels for key regions to clarify navigation cues. 
The MapNav agent uses the constructed ASM as input, harnessing VLM's powerful end-to-end capabilities to enhance vision-and-language navigation (VLN). The framework overview of MapNav is shown in Fig.~\ref{fig2}.

\subsection{Task Definition}
Vision-and-language Navigation in continuous environments (VLN-CE) refers to a task in which an agent navigates continuous 3D environments using natural language instructions as guidance.
The input to a VLN-CE agent comprises two key components: (1) a natural language instruction $\mathcal{I}$ that specifies the desired navigation path (\textit{e.g.}, ``Walk down the hallway, turn right at the plant, and stop at the third door on your left"), and (2) a sequence of first-person RGB observations ${\mathbf{X}_t}$ collected as the agent navigates through the environment.
At each timestep $t$, the agent must predict a continuous action $\mathbf{a}_{t+1} \in \mathcal{A}$ that defines its next movement, where $\mathcal{A}$ denotes the low-level action repertoire.
The agent must execute these actions sequentially until it determines that it has reached the target destination specified in the instruction.

\subsection{Our MapNav Agent}
Our MapNav agent uses the current RGB observation, Annotated Semantic Maps (ASM), and instruction to directly generate executable navigation actions. 
Specifically, we first extract the object mask from the current observation ${\mathbf{X}_t}$ using a semantic segmentation module, then construct the semantic map by integrating the depth image and pose.
To enhance readability for the VLM, we generate Annotated Semantic Maps. 
We then employ a two-stream encoder to process the observation and ASM features separately, followed by a multi-modal projector to align these modalities in a shared embedding space.
Finally, we concatenate the instruction tokens with the aligned features and input them into the VLM, which outputs executable navigation actions in text format.
The detailed architectural designs are described as follows.

\noindent \textbf{Annotated Semantic Maps (ASM) Generation.}
The generation process of our ASM is shown in Fig.~\ref{fig3}. We employ a semantic mapping system to create a rich environmental representation using a multi-channel tensor $\mathbf{M}$ of dimensions $C \times W \times H$, where $C = C_n + 4$ and $n$ represents distinct object categories. The foundational channels (1-4) encode navigational information: physical obstacles, explored regions, the agent's current position, and historical locations, while the remaining $n$ channels store object-specific semantic information.
At the start of each navigation episode, we initialize a new semantic map with the agent at $(\frac{W}{2}, \frac{H}{2})$. The map is constructed by transforming RGB-D data into point clouds, which are projected onto a 2D plane for a top-down view. By aligning semantic segmentation masks with point cloud data, we achieve accurate object-wise mapping in the dedicated channels.

\begin{figure}[!t]
\centering
\includegraphics[width=0.48\textwidth]{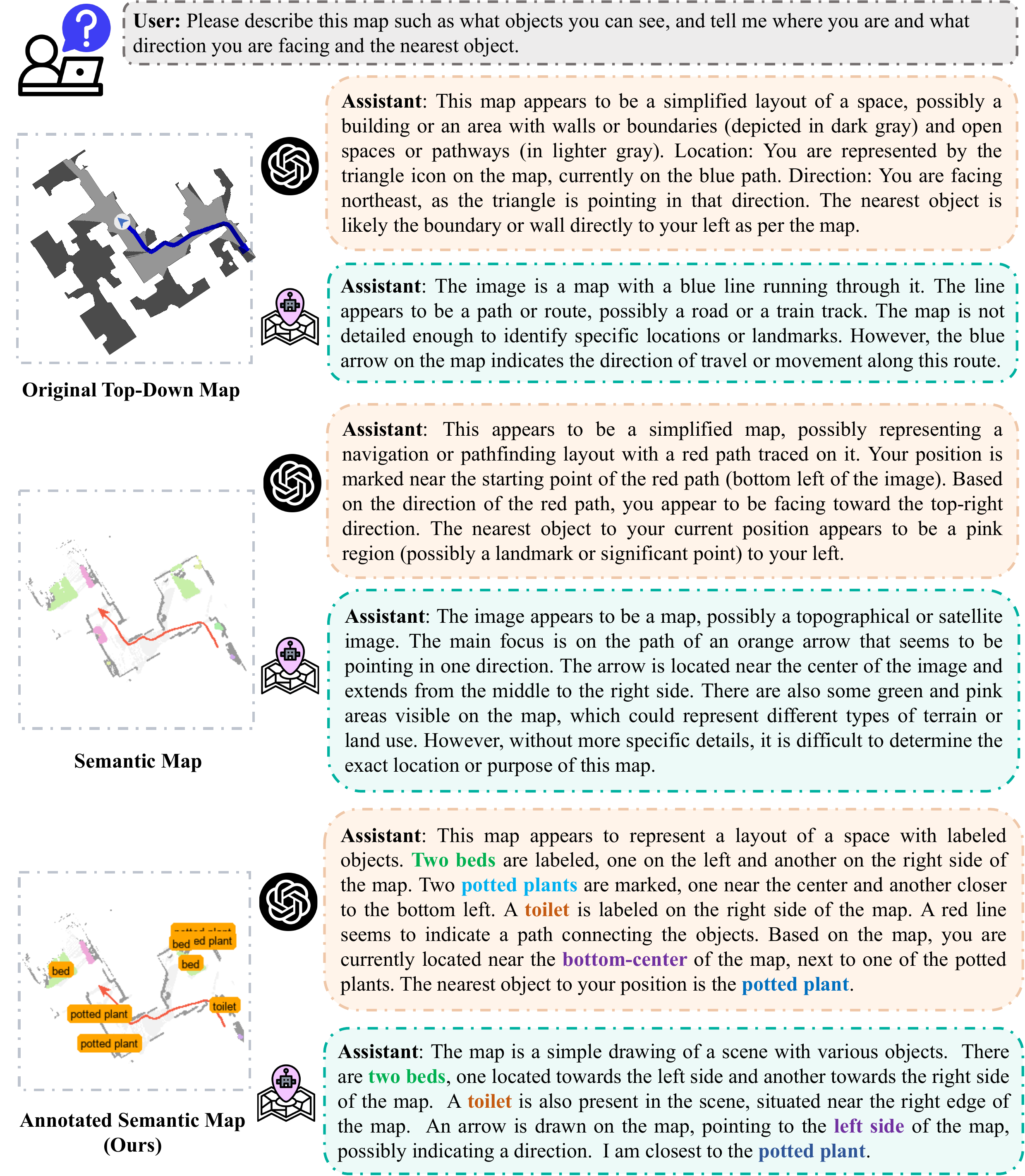}
\caption{Comparison of different VLM's understanding of different map formats includes top-down, semantic map and our ASM.}
\vspace{-15pt}
\label{fig5}
\end{figure}

\vspace{-2pt}
\noindent\textit{Remarks}: 
A key innovation in our approach is the generation of ASM following semantic map construction, enhancing traditional representations with explicit natural language annotations. The ASM generation pipeline involves two main stages: (1) semantic region identification via connected component analysis on each object-specific channel, and (2) centroid computation for these regions to determine optimal text placement.
For each semantic region exceeding the minimum area threshold $\tau$, we compute its geometric centroid to establish a text anchor point, ensuring optimal annotation placement and readability while preserving visual clarity.
As shown in Fig.~\ref{fig3}, the resulting ASM transforms abstract semantic representations into linguistically grounded spatial information (\textit{e.g.}, ``chair'', ``plant'', ``bed''), bridging spatial understanding and natural language comprehension. This explicit textual grounding allows the VLM to leverage its pre-trained knowledge of object-language relationships, facilitating intuitive spatial reasoning and improved navigation decision-making.
As shown in Fig.~\ref{fig5}, we conducted a comparative analysis of the proposed ASM against conventional top-down and semantic maps in simulator. To evaluate comprehension, we input these three map representations into both GPT-4o and MapNav. 
The results clearly demonstrate that the VLM exhibits superior understanding when processing ASM compared to traditional mapping approaches, validating our linguistic augmentation strategy.
The ASM generation process fundamentally transforms the traditional semantic mapping paradigm by introducing a novel layer of linguistic information that aligns with the VLM's pre-trained capabilities, enabling more effective multi-modal reasoning in navigation tasks. Appendix~\ref{attn} demonstrates that our ASM successfully directs the attention of the VLM towards the textual labels.

\noindent \textbf{Inputs Encoding.} 
MapNav utilizes an advanced vision-language architecture based on LLaVA-Onevision~\cite{li2024llava}. At each timestep $t$, we process two primary visual inputs: the observation frame $\mathbf{X}_t$ and the ASM $\mathbf{M}_t$. These inputs are concurrently handled by a shared vision encoder, followed by modality-specific projectors for optimal feature alignment.
Employing a SigLIP~\cite{zhai2023sigmoid} vision encoder, we derive initial visual embeddings for both inputs. 
The observation frame produces $\mathbf{X}_t \in \mathbb{R}^{N \times D}$, where $N$ represents the number of image patches and $D$ is the model's hidden dimension.
Similarly, the ASM generates $\mathbf{X}^M_t \in \mathbb{R}^{N \times D}$. Both visual features undergo parallel spatial-aware transformations through a patch merge operation:
\vspace{-2pt}
\begin{align}\label{eq:patch_merge}
&\mathbf{F}_t = \Phi_{spatial}(\mathbf{X}_t, \mathcal{G}), \notag \\
&\mathbf{F}^M_t = \Phi_{spatial}(\mathbf{X}^M_t, \mathcal{G}),
\vspace{-5pt}
\end{align}
where $\Phi_{spatial}$ denotes our spatial unpadding patch merge function and $\mathcal{G}$ defines the grid pinpoints for feature extraction. The features are subsequently projected into the language model's embedding space using modality-specific MLP projector:
\begin{equation}\label{eq:proj}
\mathbf{E}_t = P^{obs}_{mlp}(\mathbf{F}t), \quad \mathbf{E}^M_t = P^{map}_{mlp}(\mathbf{F}^M_t),
\end{equation}
where $\mathbf{E}_t, \mathbf{E}^M_t \in \mathbb{R}^{N \times C}$, with $C$ as the language model's hidden dimension. The projections $P^{obs}_{mlp}$ and $P^{map}_{mlp}$ are two-layer MLPs with GELU activation. The final multimodal representation is formed by concatenating the encoded observations and map features, along with special tokens for task structuring:
\vspace{-2pt}
\begin{equation}\label{eq:final_rep}
\mathbf{V}_t = [\text{TASK}; \mathbf{E}_t; \text{OBS}; \mathbf{E}^M_t; \text{MAP}].
\vspace{-5pt}
\end{equation}
This unified representation seamlessly integrates multiple input modalities while preserving essential spatial relationships for navigation. 
By employing efficient token management and precision optimization, the encoding process ensures computational efficiency while maximizing the model's capacity to represent complex navigation scenarios.

\begin{table*}[!t]
\small
\centering
\caption{Comparison with state-of-the-art methods on the Val-Unseen split of R2R-CE and RxR-CE. $^{*}$ indicates methods using the waypoint predictor. \textit{Cur.~RGB} and \textit{His.~RGB} refer to methods utilize current and historical RGB frames, respectively. $\dag$ indicates that we reproduced the results using the open-source code. \textit{Pano.} represents methods utilize panoramic views.}
\vspace{-10pt}
\setlength{\tabcolsep}{2pt}
\resizebox{0.95\textwidth}{!}{%
\begin{tabular}{l|ccccc|lcccc|lcccc}
\toprule
& \multicolumn{5}{c|}{Observation} & \multicolumn{5}{c|}{R2R Val-Unseen} & \multicolumn{4}{c}{RxR Val-Unseen} \\
\cmidrule{2-15}
Method & Cur.~RGB & Pano. & Depth & Odo. & His.~RGB & & NE $\downarrow$ & OS $\uparrow$ & SR $\uparrow$ & SPL $\uparrow$ & NE $\downarrow$ & SR $\uparrow$ & SPL $\uparrow$ & nDTW $\uparrow$ \\
\midrule
HPN+DN$^{*}$ & & \checkmark & \checkmark & \checkmark & \checkmark & & 6.31 & 40.0 & 36.0 & 34.0 & - & - & - & - \\
CMA$^{*}$ & & \checkmark & \checkmark & \checkmark & \checkmark & & 6.20 & 52.0 & 41.0 & 36.0 & 8.76 & 26.5 & 22.1 & 47.0 \\
VLN$\circlearrowright$BERT$^{*}$ & & \checkmark & \checkmark & \checkmark & \checkmark & & 5.74 & 53.0 & 44.0 & 39.0 & 8.98 & 27.0 & 22.6 & 46.7 \\
Sim2Sim$^{*}$ & & \checkmark & \checkmark & \checkmark & \checkmark & & 6.07 & 52.0 & 43.0 & 36.0 & - & - & - & - \\
GridMM$^{*}$ & & \checkmark & \checkmark & \checkmark & \checkmark & & 5.11 & 61.0 & 49.0 & 41.0 & - & - & - & - \\
Ego$^{2}$-Map$^{*}$ & & \checkmark & \checkmark & \checkmark & \checkmark & & 5.54 & 56.0 & 47.0 & 41.0 & - & - & - & - \\
DreamWalker$^{*}$ & & \checkmark & \checkmark & \checkmark & \checkmark & & 5.53 & 59.0 & 49.0 & 44.0 & - & - & - & - \\
Reborn$^{*}$ & & \checkmark & \checkmark & \checkmark & \checkmark & & 5.40 & 57.0 & 50.0 & 46.0 & 5.98 & 48.6 & 42.0 & 63.3 \\
ETPNav$^{*}$ & & \checkmark & \checkmark & \checkmark & \checkmark & & 4.71 & 65.0 & 57.0 & 49.0 & 5.64 & 54.7 & 44.8 & 61.9 \\
HNR$^{*}$ & & \checkmark & \checkmark & \checkmark & \checkmark & & 4.42 & 67.0 & 61.0 & 51.0 & 5.50 & 56.3 & 46.7 & 63.5 \\
BEVBert$^{*}$ & & \checkmark & \checkmark & \checkmark & \checkmark & & 4.57 & 67.0 & 59.0 & 50.0 & 4.00 & 68.5 & - & 69.6 \\
HAMT+ScaleVLN$^{*}$ & & \checkmark & \checkmark & \checkmark & \checkmark & & 4.80 & - & 55.0 & 51.0 & - & - & - & - \\
\midrule
AG-CMTP & & \checkmark & \checkmark & \checkmark & \checkmark & & 7.90 & 39.0 & 23.0 & 19.0 & - & - & - & - \\
R2R-CMTP & & \checkmark & \checkmark & \checkmark & \checkmark & & 7.90 & 38.0 & 26.0 & 22.0 & - & - & - & - \\
LAW & \checkmark & & \checkmark & \checkmark & \checkmark & & 6.83 & 44.0 & 35.0 & 31.0 & 10.90 & 8.0 & 8.0 & 38.0 \\
CM2 & \checkmark & & \checkmark & \checkmark & \checkmark & & 7.02 & 41.0 & 34.0 & 27.0 & - & - & - & - \\
WS-MGMap & \checkmark & & \checkmark & \checkmark & \checkmark & & 6.28 & 47.0 & 38.0 & 34.0 & - & - & - & - \\
AO-Planner & & \checkmark & \checkmark & & \checkmark & & 5.55 & 59.0 & 47.0 & 33.0 & 7.06 & 43.3 & 30.5 & 50.1 \\
Seq2Seq & \checkmark & & \checkmark & & \checkmark & & 7.77 & 37.0 & 25.0 & 22.0 & 12.10 & 13.9 & 11.9 & 30.8 \\
CMA & \checkmark & & \checkmark & & \checkmark & & 7.37 & 40.0 & 32.0 & 30.0 & - & - & - & - \\
RGB-Seq2Seq & \checkmark & & & & \checkmark & & 10.10 & 8.0 & 0.0 & 0.0 & - & - & - & - \\
RGB-CMA & \checkmark & & & & \checkmark & & 9.55 & 10.0 & 5.0 & 4.0 & - & - & - & - \\
\midrule
NaVid (All RGB Frames) & \checkmark & & & & \checkmark & & 5.47 & 49.1 & 37.4 & 35.9 & 8.41 & 23.8 & 21.2 & - \\
NaVid (Cur.~RGB)$\dag$ & \checkmark & & & & & & 8.10 & 24.9 & 13.0 & 7.8 & 11.33 & 8.7 & 6.8 & - \\
\rowcolor{blue!10}\textbf{MapNav(w/o ASM + Cur.~RGB)} & \checkmark & & & & & & \textbf{7.26} & \textbf{41.2} & \textbf{27.1} & \textbf{23.5} & \textbf{9.31} & \textbf{15.6} & \textbf{12.2} & \textbf{30.9} \\
\rowcolor{blue!10}\textbf{MapNav(w/ ASM + Cur.~RGB)} & \checkmark & & & & & & \textbf{5.92} & \textbf{50.3} & \textbf{36.5} & \textbf{34.3} & \textbf{8.95} & \textbf{22.1} & \textbf{20.2} & \textbf{35.6} \\
\rowcolor{blue!10}\textbf{MapNav(w/ ASM + Cur.~RGB + 2 His.~RGB)} & \checkmark & & & & \checkmark & & \textbf{5.43} & \textbf{53.0} & \textbf{39.7} & \textbf{37.2} & \textbf{7.62} & \textbf{32.6} & \textbf{27.7} & \textbf{43.5} \\
\bottomrule
\end{tabular}
}
\label{tab:r2r_rxr}
\vspace{-5pt}
\end{table*}
\begin{table*}[!t]
\centering
\caption{Comparison in diverse real-world environments scenes (\textit{Meeting Room}, \textit{Office}, \textit{Lecture Hall}, \textit{Tea Room}, and \textit{Living Room}). Simple I.F. and Semantic I.F. indicate simple and semantic instruction following tasks, respectively. Our MapNav outperforms all the baselines in both simple instructions and semantic instructions.}
\vspace{-10pt}
\setlength{\tabcolsep}{1.5pt}
\resizebox{\textwidth}{!}{%
\begin{tabular}{l|cccc|cccc|cccc|cccc|cccc}
\toprule
& \multicolumn{4}{c|}{\texttt{Meeting Room}} & \multicolumn{4}{c|}{\texttt{Office}} & \multicolumn{4}{c|}{\texttt{Lecture Hall}} & \multicolumn{4}{c|}{\texttt{Tea Room}} & \multicolumn{4}{c}{\texttt{Living Room}} \\
\cmidrule{2-21}
& \multicolumn{2}{c|}{Simple I.F.} & \multicolumn{2}{c|}{Semantic I.F.} & \multicolumn{2}{c|}{Simple I.F.} & \multicolumn{2}{c|}{Semantic I.F.} & \multicolumn{2}{c|}{Simple I.F.} & \multicolumn{2}{c|}{Semantic I.F.} & \multicolumn{2}{c|}{Simple I.F.} & \multicolumn{2}{c|}{Semantic I.F.} & \multicolumn{2}{c|}{Simple I.F.} & \multicolumn{2}{c}{Semantic I.F.} \\
\cmidrule{2-21}
Method & SR$\uparrow$ & NE$\downarrow$ & SR$\uparrow$ & NE$\downarrow$ & SR$\uparrow$ & NE$\downarrow$ & SR$\uparrow$ & NE$\downarrow$ & SR$\uparrow$ & NE$\downarrow$ & SR$\uparrow$ & NE$\downarrow$ & SR$\uparrow$ & NE$\downarrow$ & SR$\uparrow$ & NE$\downarrow$ & SR$\uparrow$ & NE$\downarrow$ & SR$\uparrow$ & NE$\downarrow$ \\
\midrule
WS-MGMap & 50\% & 1.62 & 20\% & 2.83 & 60\% & 1.21 & 30\% & 3.11 & 60\% & 0.95 & 20\% & 3.63 & 50\% & 1.11 & 20\% & 2.93 & 70\% & 0.62 & 40\% & 2.86 \\
Navid & 70\% & 0.86 & 50\% & 1.93 & 70\% & 1.59 & 60\% & 1.99 & 60\% & 0.75 & 40\% & 2.94 & 80\% & 0.35 & 50\% & 1.61 & 60\% & 2.13 & 30\% & 3.83 \\
\rowcolor{blue!10}\textbf{MapNav(Ours)} & \textbf{70\%} & \textbf{0.73} & \textbf{60\%} & \textbf{1.32} & \textbf{80\%} & \textbf{0.96} & \textbf{60\%} & \textbf{1.38} & \textbf{80\%} & \textbf{0.82} & \textbf{70\%} & \textbf{1.15} & \textbf{90\%} & \textbf{0.31} & \textbf{70\%} & \textbf{0.66} & \textbf{80\%} & \textbf{1.03} & \textbf{60\%} & \textbf{0.85} \\
\bottomrule
\end{tabular}
}
\label{tab:real}
\vspace{-10pt}
\end{table*}
\noindent \textbf{Action Prediction.} 
Our framework implements a fully end-to-end Action Prediction strategy that directly generates navigation commands in natural language form. Unlike traditional approaches that require separate action decoders, we leverage the VLM's inherent language understanding capabilities to directly parse navigation intentions into discrete actions. 
 
To robustly interpret the model's natural language outputs, we employ a comprehensive pattern matching system capable of recognizing various linguistic expressions of the same navigational intent:

\vspace{-8pt}
\begin{equation}\label{eq:action_parse}
\mathcal{A}(t) = \Psi(\mathcal{T}(t), \mathcal{P}),
\vspace{-8pt}
\end{equation}
where $\mathcal{T}(t)$ denotes the model's text output at time $t$, $\mathcal{P}$ represents our pattern matching ruleset, and $\Psi$ is the action parsing function that translates natural language into discrete actions. For each action type, we maintain a comprehensive collection of synonymous expressions:

\vspace{-5pt}
\begin{align*}
&\mathcal{P}_{\textit{FORWARD}} = {\textit{\{``move forward''}, \textit{``proceed''}, ...\}}, \\
&\mathcal{P}_{\textit{TURN-LEFT}} = {\textit{\{``turn left''}, \textit{``rotate left''}, ...\}}, \\
&\mathcal{P}_{\textit{TURN-RIGHT}} = {\textit{\{``turn right''}, \textit{``rotate right''}, ...\}}, \\
&\mathcal{P}_{\textit{STOP}} = {\textit{\{``stop''}, \textit{``halt''}, \textit{``wait''}, ...\}}.
\vspace{-13pt}
\end{align*}
This natural language interpretation approach presents several advantages. First, it eliminates the need for additional decoder networks, thereby simplifying the architecture. Second, it preserves the end-to-end nature of the system, enhancing efficiency. Finally, it offers robustness against variations in language expression.
Pattern matching is achieved through case-insensitive regular expressions that accommodate various word arrangements, which enhances the system's resilience to the inherent variations in VLM's output formatting.

\section{Experiments}
\label{sec5:experiments}
\subsection{Experimental Details}
\noindent\textbf{Dataset.}
We constructed a comprehensive dataset of approximately 1 Million training pairs using a hybrid collection strategy, which includes ground truth trajectories from the R2R and RxR datasets ($\approx$300k pairs from both), DAgger-collected data ($\approx$200k pairs from both), and specialized collision recovery samples ($\approx$25k pairs from both). 
This approach ensures diverse coverage of navigation scenarios and incorporates samples from general vision-language tasks for co-training.

For fair comparison, we trained on 500k pairs from R2R and evaluated Zero-Shot capability on RxR, in addition to conducting separate training and evaluation on the RxR dataset. 
Detailed strategies for dataset construction are provided in Appendix~\ref{sec3:dataset}.

\noindent\textbf{Simulated Environments.}
We evaluate our \textbf{\textit{MapNav}} agent using the VLN-CE benchmark in Habitat, which offers a continuous environment for navigation in reconstructed, photo-realistic indoor scenes.
We focus on the val-unseen split of the R2R and RxR datasets in VLN-CE, which are two of the most recognized benchmarks in VLN, comprising 1,839 and 1,517 episodes, respectively.

\noindent\textbf{Real-world environments. } 
To evaluate the sim-to-real performance of our model, we designed diverse real-world experiments featuring 50 instructions across five environments: office, conference room, lecture hall, living room, and tea room. Simple instructions require basic actions like moving forward and turning, while semantic instructions involve navigating to specific objects (\textit{e.g.}, ``move forward to the refrigerator and then turn right'').

\noindent\textbf{Metrics.}
We utilize several widely used  evaluation metrics for VLN tasks: Navigation Error (NE), Oracle Success rate (OS), Success Rate (SR), Success weighted Path Length (SPL), and normalized Dynamic Temporal Wrapping (nDTW). 
\textbf{\textit{SPL}} is the primary metric, as it effectively reflects both navigation accuracy and efficiency.

\noindent\textbf{Implementation Details.}
Our model is trained on a cluster server with 8 NVIDIA A100 GPUs for about 30 hours, totaling 240 GPU hours. We employ the LLaVA-Onevision~\cite{li2024llava} architecture, using Google's SigLIP-so400M~\cite{zhai2023sigmoid} as the vision encoder, Qwen2-7B-Instruct~\cite{wang2024qwen2} as the language model backbone, and Mask2Former~\cite{cheng2021mask2former} for semantic segmentation during ASM generation. 
For detailed experimental details, see Appendix~\ref{sec3:training_details}.

\subsection{Comparisons with SOTA Methods}
\noindent\textbf{Results in Simulated Environment.}
To evaluate cross-dataset performance, we firstly train solely on R2R samples and then assess its zero-shot performance on the RxR Val-Unseen split.
To ensure fairness and consistency in our comparison, we first conducted experiments on single-frame RGB, where \textit{\textbf{MapNav (w/o ASM + Cur.~RGB)}} denotes MapNav without ASM, and NaVid (Cur.~RGB) refers to NaVid without historical frames. As shown in Tab.~\ref{tab:r2r_rxr}, we achieved improvements of 14.1\% in SR and 15.7\% in SPL for R2R, and 6.9\% in SR and 5.4\% in SPL for RxR.
After incorporating ASM, \textit{\textbf{MapNav (w/ ASM + Cur.~RGB)}}, which uses single-frame RGB images and ASM as input, shows improvements of 23.5\% in SR and 26.5\% in SPL for R2R compared to Navid (Cur.~RGB). 

Furthermore, our performance is competitive with Navid (All Frames), demonstrating that ASM effectively serves as a new historical representation method for large models, replacing traditional historical frames.

After incorporating just two historical RGB frames and training on both R2R and RxR, \textit{\textbf{MapNav (w/ ASM + Cur.~RGB + 2 His.RGB)}} outperforms state-of-the-art methods that use all historical frames. Specifically, compared to the previous SOTA method, Navid~(All RGB Frames), we achieved improvements of 2.5\% in SR and 1.3\% in SPL for R2R, as well as 8.8\% in SR and 6.5\% in SPL for RxR.

\begin{table}[!t]
\centering
\caption{Memory consumption and average processing time comparison between Navid and MapNav (our method) across different numbers of navigation steps.}
\resizebox{\linewidth}{!}{
\scalebox{1}{
\setlength{\tabcolsep}{2.5mm}{
\begin{tabular}{l|cccc|c}
\hline
\multirow{2}{*}{Method} & \multicolumn{4}{c|}{Memory Consumption$\downarrow$} & \multirow{2}{*}{Avg. Time$\downarrow$} \\ \cline{2-5}
& 1 Step & 10 Steps & 100 Steps & 300 Steps & \\
\hline \hline
Navid & 0.92MB & 9.2 MB & 92 MB & 276 MB & 1.22s \\
\rowcolor{blue!10}\textbf{MapNav~(Ours)} & \textbf{0.17MB} & \textbf{0.17MB} & \textbf{0.17MB} & \textbf{0.17MB} & \textbf{0.25s} \\
\hline
\end{tabular}
    }
}}
\label{tab:inference}
\vspace{-1.0em}
\end{table}

\noindent\textbf{Results in Real-World Environment.} 
We utilize two metrics (\textit{SR} and \textit{NE}) to compare \textbf{\textit{MapNav(w/ ASM + Cur.~RGB)}} with WS-MGMAP~\cite{chen2022weakly} and Navid~\cite{zhang2024navid}, which both use all the historical frames. 
As shown in Tab.~\ref{tab:real}, MapNav significantly outperforms both WS-MGMAP and Navid in SR and NE across simple and semantic instructions. 
Specifically, our method surpasses Navid under the semantic instruction settings in the lecture hall and living room, where we improve the \textit{SR} by 30\% in each setting.
These results highlight the exceptional performance of \textbf{\textit{MapNav}} in real-world scenarios and validate the effectiveness of our proposed ASM.

\subsection{Ablation Studies}

\noindent\textbf{Efficiency Analysis.}
For the efficiency analysis, we compare memory consumption and inference time between MapNav and Navid.
As shown in Tab.~\ref{tab:inference}, MapNav demonstrates significant improvements in both areas.
Specifically, our semantic map-based method maintains a constant memory footprint of 0.17MB, regardless of trajectory length, while Navid's frame-based approach scales linearly, reaching 276MB at 300 steps. This is because MapNav only stores and updates a compact ASM, whereas Navid accumulates all historical RGB observations.
In terms of inference speed, \textbf{\textit{MapNav}} reduces average processing time by 79.5\% (from 1.22 seconds per step to 0.25 seconds). This improvement arises because MapNav only calculates features from the current RGB frame and ASM, avoiding the need to process all historical frames.

\noindent\textbf{Effect of ASM.}
To validate the efficacy of our proposed ASM, we conducted a comparative analysis using multiple map variants. We evaluated three conditions: (1) the original top-down map from the simulator, (2) a semantic map with categorical information but no textual annotations, and (3) complete removal of map-based features.
Quantitative results in Tab.~\ref{tab:ablationmap} show a clear performance hierarchy: the RGB-only baseline performed the worst, with moderate improvements from the original and semantic map variants. The ASM outperformed all evaluated metrics, highlighting the benefits of semantic enrichment and the importance of textual annotations for spatial reasoning tasks.

\begin{table}\centering
\caption{
Comparison of different map representation methods.}

\resizebox{\linewidth}{!}{
\scalebox{1}{
\setlength{\tabcolsep}{2.5mm}{
\begin{tabular}{l|cccc}
\hline
 & \multicolumn{4}{c}{VLN-CE R2R Val-Unseen} \\ \cline{2-5} 
Method & NE↓ & OS↑ & SR↑ & SPL↑ \\ \hline \hline
MapNav (w/o Map)  & 7.26 & 41.2 & 27.3 & 23.2 \\
MapNav (Original Map) & 8.93 & 35.1 & 26.4 & 21.9 \\
MapNav (Semantic Map)  & 6.56 & 43.2 & 29.1 & 24.5 \\ \hline
 \rowcolor{blue!10} \textbf{MapNav~(ASM)}  & \textbf{5.22} & \textbf{50.3} & \textbf{36.5} & \textbf{34.3} \\ \hline
\end{tabular}
    }
}}
\vspace{-7pt}
\label{tab:ablationmap}
\end{table} 

\begin{table}[t]\centering
\caption{
Comparison of different training dataset composition.}

\vspace{-2pt}
\resizebox{\linewidth}{!}{
\scalebox{1}{
\setlength{\tabcolsep}{2.5mm}{
\begin{tabular}{l|cccc}
\hline
 & \multicolumn{4}{c}{VLN-CE R2R Val-Unseen} \\ \cline{2-5} 
Method  & NE↓ & OS↑ & SR↑ & SPL↑ \\ \hline \hline
MapNav (300k)  & 6.38 & 38.2 & 23.9 & 19.5 \\
MapNav (300k+DAgger)  & 6.02 & 46.1 & 33.5 & 30.7 \\  
MapNav (300k+DAgger+RxR)  & 5.89 & 48.2 & 34.4 & 31.7 \\ \hline

 \rowcolor{blue!10} \textbf{MapNav (300k+DAgger+Collision)} & \textbf{5.22} & \textbf{50.3} & \textbf{36.5} & \textbf{34.3} \\ \hline
\end{tabular}
    }
}}
\vspace{-10pt}
\label{tab:ablationdataset}
\end{table}

\noindent\textbf{Impact of different training dataset composition.}
To evaluate the impact of different training data compositions, we conducted a dataset ablation study with five configurations: (1) MapNav trained on 100K R2R samples (baseline), (2) 300K R2R samples, (3) DAgger-generated samples on R2R, (4) integration of DAgger and RxR datasets, and (5) all previous components plus collision-aware training.
As shown in Tab.~\ref{tab:ablationdataset}, increasing from 100K to 300K R2R samples resulted in substantial performance improvements. The integration of DAgger samples yielded the most significant gains, underscoring the importance of interactive learning. The RxR dataset provided modest improvements, particularly in diverse linguistic instructions.
Our final configuration, with collision-aware training, achieved marginally better performance across metrics, setting a new performance benchmark. These results highlight that data diversity and interactive learning enhance model performance, with the greatest benefits stemming from ample base training data and DAgger augmentation.

\begin{figure}[!t]
\centering
\includegraphics[width=0.45\textwidth]{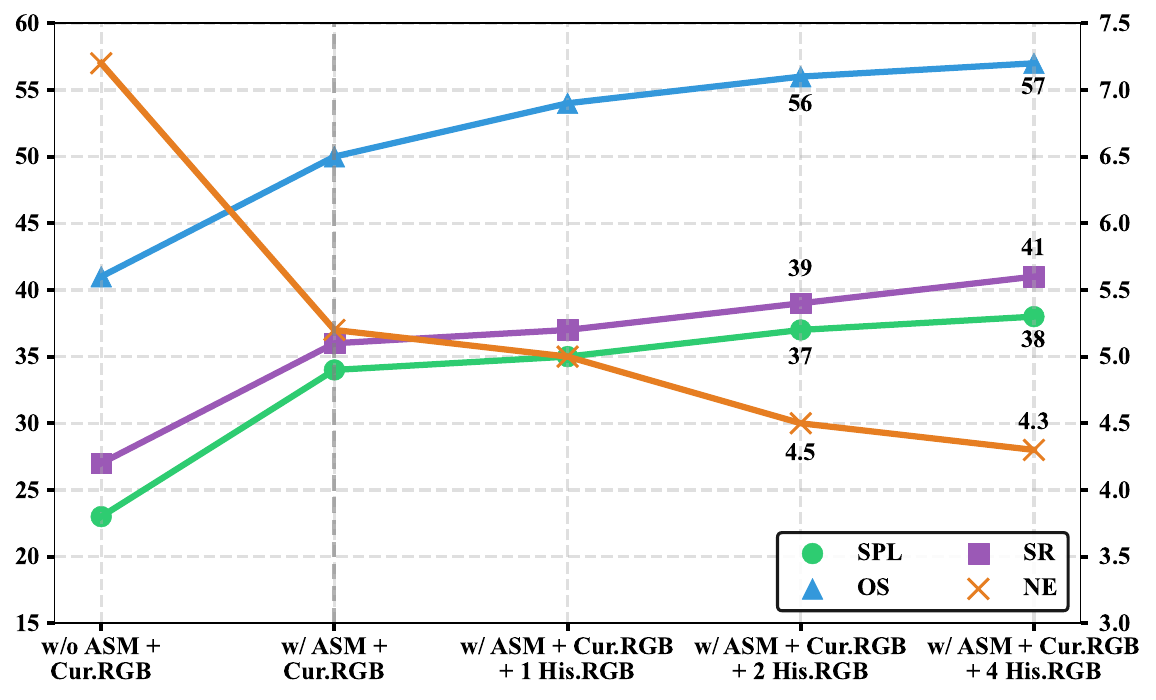}
\vspace{-5pt}
\caption{Comparison of MapNav using different numbers of historical RGB frames. Cur.~RGB and His.~RGB refer to methods using the current and historical RGB frames, respectively.}
\vspace{-15pt}
\label{fig_his}
\end{figure}

\noindent\textbf{Effect of Varying Numbers of Historical RGB Frames.}
We conducted a systematic ablation study to evaluate the effect of varying historical RGB frames. As shown in Fig.~\ref{fig_his}, the most significant improvement occurs with ASM in the single current RGB frame setting, enhancing SR from 27\% to 36\%, SPL from 23\% to 34\%, OS from 41\% to 50\%, and NE from 7.2m to 5.2m. This highlights the effectiveness of our ASM-based approach in capturing spatial information.
While adding historical RGB frames (1, 2, and 4 His.RGB) results in gradual improvements, these gains are modest compared to the initial ASM integration. Notably, with just two historical RGB frames, our model achieves state-of-the-art performance (SR: 39\%, SPL: 37\%) compared to the baseline NaVid~\cite{zhang2024navid}.
The final configuration with four historical RGB frames shows only slight gains (SR: 41\%, SPL: 38\%), indicating that the core advantage lies in the ASM-based representation rather than temporal accumulation. These findings validate our choice of ASM as a more efficient alternative to extensive historical RGB frame sequences, delivering superior performance while maintaining computational efficiency.

\section{Conclusion}
\label{sec6:conclusion}
In this paper, we introduce MapNav, a novel method using Annotated Semantic Maps (ASMs) for Vision-and-Language Navigation. It achieves state-of-the-art performance while significantly reducing memory usage compared to traditional approaches. By converting RGB-D and pose data into enriched top-down maps with annotations, MapNav integrates efficiently with Vision-Language Models. Experiments show it delivers superior navigation while maintaining a constant memory footprint of 0.17MB, regardless of trajectory length.

\section*{Limitations}
While MapNav demonstrates promising results, the semantic segmentation module may occasionally produce imprecise object labels under challenging conditions like occlusions or varying lighting. To address this aspect, we have identified several promising directions for future research. First, we aim to explore more advanced semantic understanding approaches that can effectively handle dynamic scenes and partial observations through multi-view scene understanding. Second, we plan to investigate methods to enhance real-world generalization by leveraging the extensive real-world image data in VLM pre-training. Additionally, we intend to extend MapNav to tackle more complex embodied AI tasks, such as interactive navigation and manipulation, which will require integrating object affordances and physical interaction capabilities into the ASM representation.

\section*{Acknowledgement}
This work was supported by the National Science and Technology Major Project (No. 2022ZD0117800).

\bibliography{references}

\newpage

\newpage
\clearpage
\appendix
\section*{Appendix}
\label{sec:appendix}

This supplementary material provides additional details on the proposed method and experimental results that could not be included in the main manuscript due to page limitations.

Specifically, this appendix is organized as follows:
\begin{itemize}
    \item Sec.~\ref{sec3:dataset} provides a comprehensive overview of the dataset construction process. 
    \item Sec.~\ref{sec3:training_details} outlines the detailed training procedures used for our models.
    \item Sec.~\ref{sec:realworldrobot} describes our real-world MapNav robot setup along with implementation specifics.
    \item Sec.~\ref{sec:Qualitative} presents the qualitative results in both the simulator and real-world environments.
    \item Sec.~\ref{sec3:more_experiments} showcases additional experimental results that further validate our findings.
    \item Sec.~\ref{sec:append_simdemo} offers further qualitative results specifically from the simulator.
    \item Sec.~\ref{sec:append_realdemo} provides additional qualitative results obtained from real-world scenarios.

    \item Sec.~\ref{attn} analyzes VLM attention visualization across different map representations.
\end{itemize}

\section{Dataset Construction}
\label{sec3:dataset}

\textbf{Motivation and Overview.}
 Training data for VLN tasks faces two primary challenges: limited diversity and insufficient scale. To address these challenges, we propose a comprehensive data collection strategy that combines three complementary approaches: expert demonstrations based on ground truth trajectories, interactive learning through DAgger, and specialized collision recovery data. This hybrid strategy maximizes the utility of the available training data and significantly enhances the model's generalization capabilities across diverse scenarios.
 
\noindent\textbf{Phase I: Expert Demonstration Collection.} 
Initially, we collected around 300k high-quality training pairs from both R2R and RxR datasets. Each step-wise pair consists of three components: the current RGB observation frame, the corresponding Annotated Semanti Map~(ASM), and the associated action. These pairs serve as the foundation for our initial supervised fine-tuning (SFT) process for the VLM, providing the model with expert demonstration data from diverse indoor environments.

\noindent\textbf{Phase II: Interactive Learning via DAgger.} 
Following the Dataset Aggregation (DAgger) methodology \cite{ross2011dagger}, we deployed our pretrained model to collect additional trajectory data, amassing approximately 200k new training pairs from both R2R and RxR environments. This phase was crucial for two reasons:
First, combining expert trajectories with DAgger-collected data creates a more robust training dataset that reflects both ideal navigation behavior and realistic agent interactions, including potential navigation errors and recovery strategies. Second, this hybrid approach bridges the gap between training and deployment conditions, enhancing the model's ability to tackle novel scenarios and unexpected situations during navigation.

\noindent\textbf{Phase III: Specialized Collision Recovery Data.}
To overcome the limitation of expert trajectories rarely including collision scenarios, we supplemented our dataset with specialized collision recovery data. During the DAgger collection strategy, we specifically gathered instances where the agent encountered obstacles and needed recovery actions. This additional dataset, consisting of approximately 25k step-wise training pairs from both R2R and RxR, captures the agent's collision recovery behavior through rotational movements and obstacle avoidance strategies. This collision recovery subset was crucial for enhancing the model's robustness in deployment scenarios. While ground truth trajectories provide optimal navigation patterns and DAgger data captures general interaction scenarios, the collision recovery dataset specifically addresses edge cases where the agent must navigate obstacles. During the final fine-tuning phase, we integrated this collision recovery data with the primary dataset, enabling the model to learn effective obstacle avoidance and recovery strategies.

\noindent\textbf{Overall Dataset Structure and Scale.}
The complete dataset composition can be formalized as:
\begin{align}\label{eq:dataset}
&\ \mathcal{D}_{total\_R2R} = \mathcal{D}_{{total}\_{RxR}}  \\ \notag
&\ = \mathcal{D}_{GT} \cup \mathcal{D}_{DAgger} \cup \mathcal{D}_{Collision},
\end{align}
where $|\mathcal{D}_{GT}|~\approx~|\mathcal{D}_{DAgger}|~\approx~200k$, and $|\mathcal{D}_{collision\_R2R}| \approx |\mathcal{D}_{collision\_RxR}| \approx 25k$ for both R2R and RxR environments respectively.

This comprehensive dataset collection strategy significantly improves the agent's ability to handle unforeseen obstacles and navigate challenging situations autonomously. The enhanced training dataset, now totaling approximately \textbf{1 Million} training pairs, offers a more complete representation of navigation scenarios, encompassing both optimal pathfinding and practical recovery strategies. This hybrid data collection strategy substantially contributes to the model's resilience and adaptability across diverse indoor environments and navigation challenges.

\noindent\textbf{Incorporating General Vision-Language Datasets.}
To enhance our model's visual understanding and general reasoning capabilities, we implement a co-training strategy that leverages both navigation-specific data and general-purpose vision-language datasets. Following the approach similar to LLaVA-OneVision~\cite{li2024llava}, we incorporate a diverse set of high-quality visual understanding datasets alongside our navigation training data.
Specifically, for R2R task, the co-training process utilizes approximately 600k samples from general vision-language tasks, which complement our 500k navigation training samples. 
The auxiliary dataset encompasses diverse visual understanding tasks including Visual Question Answering (VQA) samples for enhanced visual reasoning, multi-image reasoning tasks for improved cross-image understanding and relationship inference, and video-based tasks for strengthening temporal reasoning and dynamic scene comprehension. For more details, please refer to~\cite{li2024llava}.

\section{Details of Training}
\label{sec3:training_details}
\textbf{Model Setting.}
Our model builds upon the LLaVA-Onevision~\cite{li2024llava} framework and consists of three main components: the visual encoder, projector, and large language model (LLM). For the visual encoder, we employed Google's SigLIP-so400m-patch14-384~\cite{zhai2023sigmoid}, which processes input images at a resolution of 384x384 using patches of size 14x14. The SigLIP model serves as our vision backbone, converting input images into visual embeddings with a hidden dimension of 1152. The projector is implemented as a two-layer MLP with GELU activation, which maps the visual features to the language model's embedding space. For the language model, we utilized Qwen2-7B-Instruct~\cite{wang2024qwen2} as our backbone, which features 28 transformer layers with 28 attention heads and a hidden size of 3584. The model supports a context length of up to 32,768 tokens and incorporates sliding window attention with a window size of 131,072 tokens.

\noindent\textbf{Training Setting.}
We conducted our training on 8 NVIDIA A100 GPUs for approximately 30 hours, totaling 240 GPU hours ($\approx$500k step-wise samples). During the fine-tuning process, we froze the vision encoder and only fine-tuned the multi-modal projector and language model components for one epoch with a learning rate of 1e-6. We utilized bfloat16 precision for training efficiency. The model processes images using bilinear spatial pooling and selects features from the penultimate layer of the vision tower. For multi-modal integration, we employed patch-level features without using image patch tokens or explicit image start/end tokens. The projector adopts a spatial unpadded patch merge strategy to handle varying image sizes effectively.

\begin{figure}[t]
\centering
\includegraphics[width=0.45\textwidth]{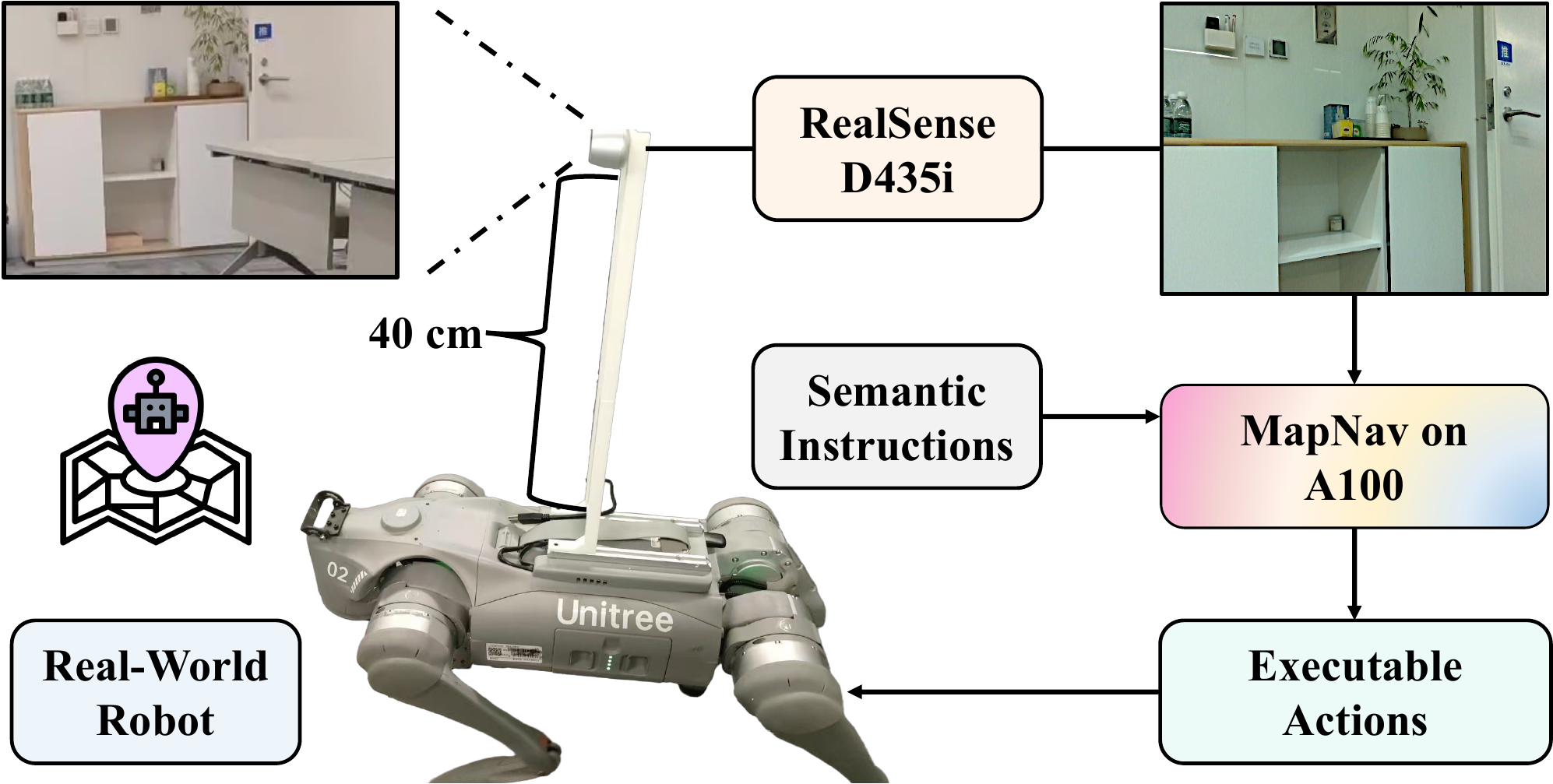}
\caption{The real-world MapNav robot setup.}
\label{figrealrobot}
\end{figure}
\section{Real-World Robot Setup}
\label{sec:realworldrobot}
As shown in Fig.~\ref{figrealrobot}, our real-world experiments were conducted using a Unitree Go2 Edu quadrupedal robot equipped with an Intel RealSense D435i depth camera mounted 40cm above the robot. The system streams RGB-D images from a D435i camera to a server powered by NVIDIA A100 GPUs, where our MapNav agent processes the observations to generate our ASMs. Based on the generated ASM, the system infers executable actions which are then transmitted back to the Go2 for execution. Notably, our system demonstrated robust sim-to-real transfer capabilities, maintaining accurate ASM generation and high task performance despite real-world challenges such as depth measurement uncertainties, pose estimation errors, and environmental variations.

\begin{figure*}[t]
\centering
\includegraphics[width=0.9\textwidth]{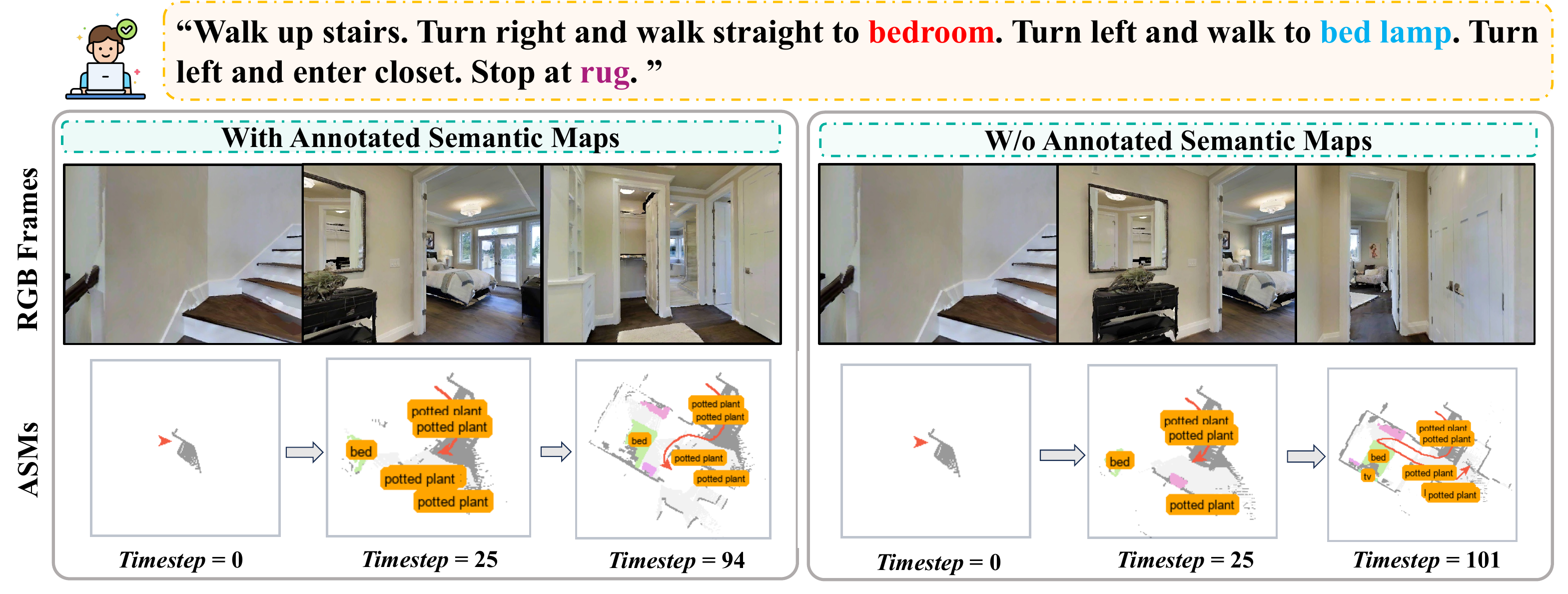}
\caption{Visualization results of \textit{\textbf{MapNav}} in the simulator.}
\label{fig:sim_demo}
\end{figure*}

\begin{figure*}[t]
\centering
\includegraphics[width=0.9\textwidth]{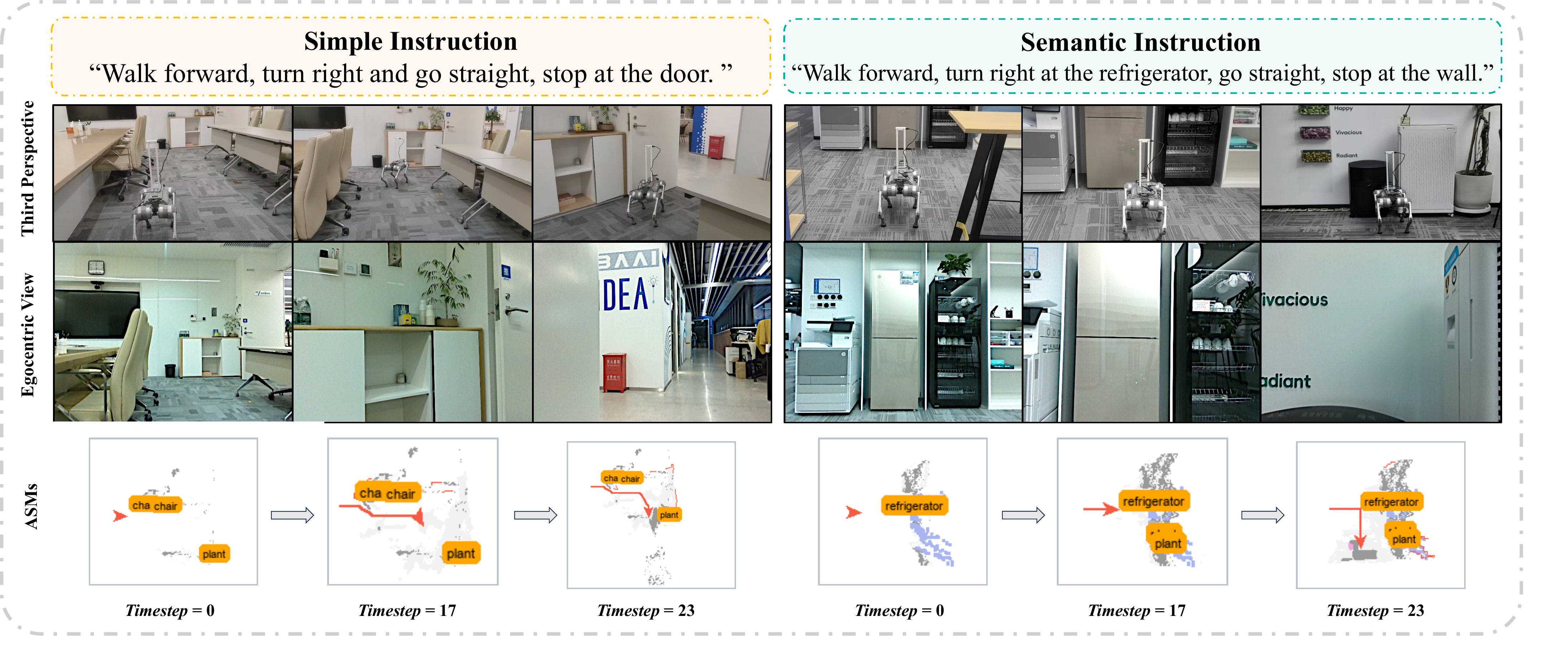}
\caption{Visualization results of \textit{\textbf{MapNav}} in the real-world.}
\vspace{-10pt}
\label{fig:real_demo}
\end{figure*}
\section{Qualitative Results.}
\label{sec:Qualitative}
Fig.~\ref{fig:sim_demo} and Fig.~\ref{fig:real_demo} illustrate the performance of our MapNav agent in simulated and real-world environments.
In the simulator, the agent with ASM successfully identifies the shortest path while following complex instructions involving multiple waypoints. In contrast, without ASM, the agent struggles to find the correct path, underscoring ASM's importance in semantic understanding and path planning.
In real-world tests, the agent effectively executes simple navigation instructions and excels at complex tasks involving semantic landmarks. The ASM visualization reveals its ability to adapt paths in real-time, confirming robust sim-to-real transfer.
These results demonstrate that our approach maintains strong performance across both environments, particularly in tasks requiring semantic understanding and adaptive navigation.

\section{More Experiments}
\label{sec3:more_experiments}

\noindent\textbf{Ablation on Different Input Modals.}
To systematically investigate the contribution of each input modality, we conducted an extensive ablation study on the input representations. The experimental configurations encompassed three variants: (1) RGB-only input, serving as the baseline condition, (2) RGB augmented with depth information (RGB+Depth), and (3) our complete model utilizing the ASM alongside RGB inputs. Quantitative results, as presented in Tab .~\ref{tab:ablationarchitecture}, reveal an interesting pattern across these configurations. While the RGB-only baseline achieved moderate performance, the addition of depth information (RGB+Depth) led to a notable performance degradation. This counterintuitive result can be attributed to the Vision-Language Model's inherent limitations in processing depth modalities, as these models are primarily trained on RGB images and natural language. In contrast, the integration of our proposed ASM demonstrated substantial performance gains across all evaluation metrics, significantly outperforming both baseline configurations. These results not only validate the effectiveness of our ASM approach but also highlight the importance of selecting input modalities that align with the pre-trained model's capabilities.


\begin{table}\centering
\caption{
Comparison of different input modals.}

\resizebox{\linewidth}{!}{
\scalebox{1}{
\setlength{\tabcolsep}{2.5mm}{
\begin{tabular}{l|cccc}
\hline
 & \multicolumn{4}{c}{VLN-CE R2R Val-Unseen} \\ \cline{2-5} 
Method  & NE↓ & OS↑ & SR↑ & SPL↑ \\ \hline \hline
MapNav (Only RGB)  & 7.26 & 41.2 & 27.1 & 23.5 \\
MapNav (RGB+Depth)  & 8.82 & 35.6 & 23.1 & 19.9 \\ \hline
\rowcolor{blue!10}\textbf{MapNav (Annotated Semantic Map)}  & \textbf{5.22} & \textbf{50.3} & \textbf{36.5} & \textbf{34.3} \\ \hline
\end{tabular}
    }
}}
\label{tab:ablationarchitecture}
\end{table} 
\begin{table}[t]\centering
\caption{
Comparison of different semantic segmentation modules' performance on R2R Val-Unseen split.
}

\resizebox{\linewidth}{!}{
\scalebox{1}{
\setlength{\tabcolsep}{2.5mm}{
\begin{tabular}{l|cccc}
\hline
 \multirow{2}{*}{VLMs}& \multicolumn{4}{c}{VLN-CE R2R Val-Unseen} \\ \cline{2-5} 
  & NE↓ & OS↑ & SR↑ & SPL↑  \\ \hline \hline
YOLOv8~\cite{Jocher_Ultralytics_YOLO_2023}  & 6.43 & 45.2 & 31.5 & 29.6 \\
MobileSAM~\cite{mobile_sam}  & 6.02 & 48.9 & 34.8 & 31.6 \\ 
\hline
\rowcolor{blue!10}\textbf{Mask2Former (Our Use)}~\cite{cheng2021mask2former}  & \textbf{5.22} & \textbf{50.3} & \textbf{36.5} & \textbf{34.3} \\ \hline
\end{tabular}
    }
}}
\label{tab:ablationsegmentation}
\end{table} 

\noindent\textbf{Ablation Experiment on Different Semantic Segmentation Modules.}
We conducted another ablation study to evaluate the impact of different semantic segmentation modules on our ASM generation process, with results presented in Tab.~\ref{tab:ablationsegmentation}. We compared three state-of-the-art semantic segmentation approaches: YOLOv8~\cite{Jocher_Ultralytics_YOLO_2023}, MobileSAM~\cite{mobile_sam}, and our chosen Mask2Former~\cite{cheng2021mask2former}. The experiments were conducted on the R2R Val-Unseen split to assess generalization capability. Results show that Mask2Former achieves the best performance across all metrics, with a Navigation Error (NE) of 5.22m, Success Rate (SR) of 36.5\%, and SPL of 34\%, outperforming both YOLOv8 and MobileSAM. The superior performance of Mask2Former can be attributed to its more precise segmentation boundaries and better handling of complex indoor scenes. While MobileSAM shows moderate performance with an SR of 34.8\% and SPL of 31.6\%, YOLOv8 exhibits relatively lower performance with an SR of 31.5\% and SPL of 29.6\%. These results suggest that the quality of semantic segmentation significantly impacts the overall navigation performance, highlighting the importance of accurate semantic objects segmentation in ASM generation.


\begin{figure*}[t]
\centering
\includegraphics[width=0.87\textwidth]{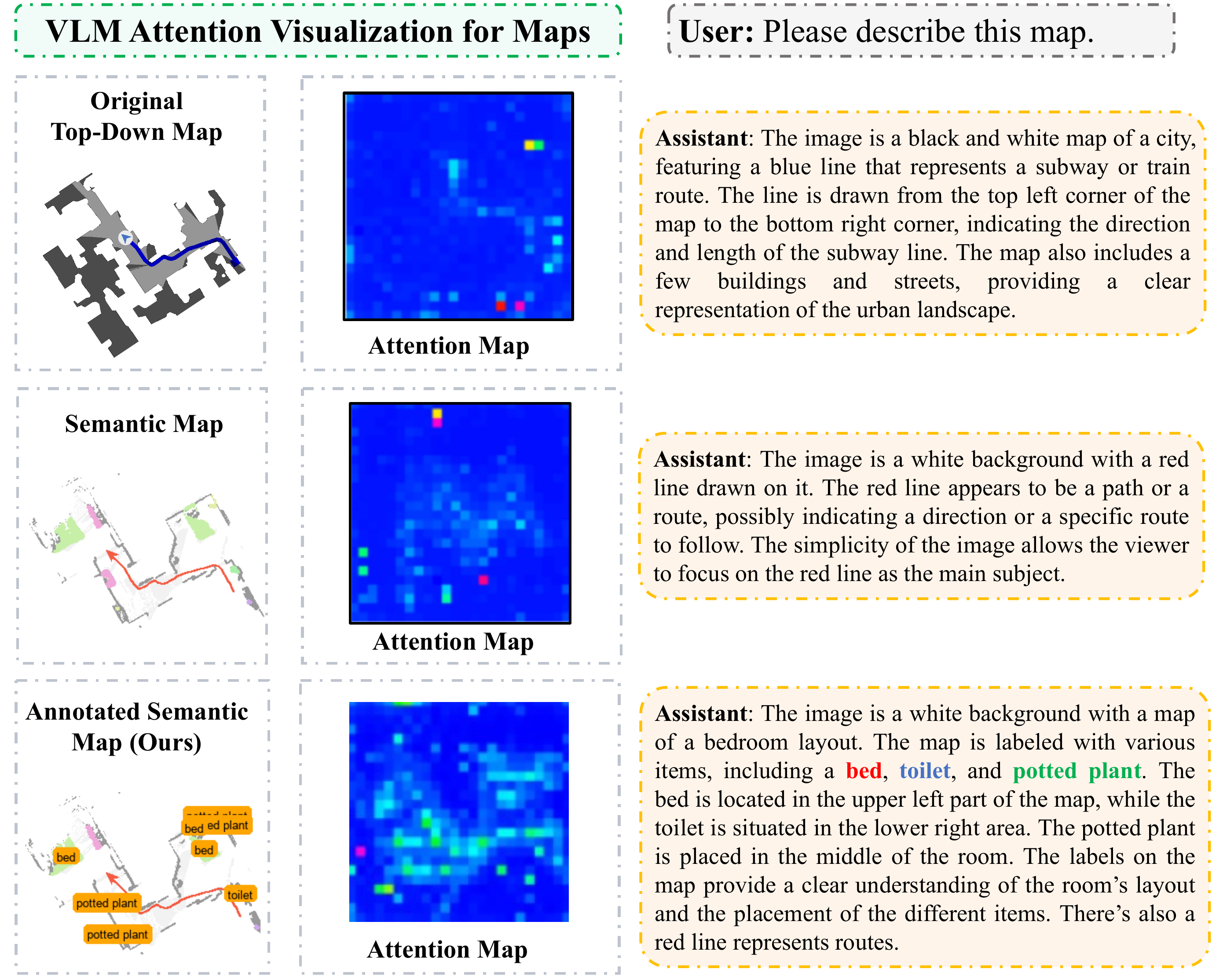}
\caption{\textbf{Visualization of VLM Attention Across Different Map Representations.} A comparison of different map representations showing that while Semantic Map exhibits sparse attention patterns without convergence on semantic objects, our ASM successfully leverages textual labels to guide attention towards semantic objects, as evidenced by concentrated attention distributions and the VLM's responses.}
\label{figattnmap}
\end{figure*}

\section{More Simulated Demos} 
\label{sec:append_simdemo}
We conducted extensive visualization experiments to demonstrate the effectiveness of our approach, selecting representative successful cases from both R2R and RxR datasets. Specifically, as shown in Fig.~\ref{fig:append_simdemo1}--Fig.~\ref{fig:append_simdemo6},  we visualized 24 distinct navigation trajectories across six pages, with each case highlighting the agent's navigation process and corresponding Abstract Semantic Maps (ASM). These visualizations showcase the robust generalization capability of our ASM-based approach across diverse environments and navigation scenarios. The demonstrated cases showcase a variety of room layouts, navigation objectives, and complex multi-step instructions, all of which our agent successfully interpreted and executed. This performance across diverse scenarios validates our approach's spatial understanding capabilities, as reflected in the agent's ability to generate accurate ASMs and perform appropriate navigation actions. Additionally, these examples highlight our ASMs' strong memory representation abilities, illustrating how the agent maintains and updates its understanding of the environment throughout the navigation process. By integrating new observations with previously stored spatial information, the agent can make informed navigation decisions.
\begin{figure*}[t]
\centering
\includegraphics[width=0.87\textwidth]{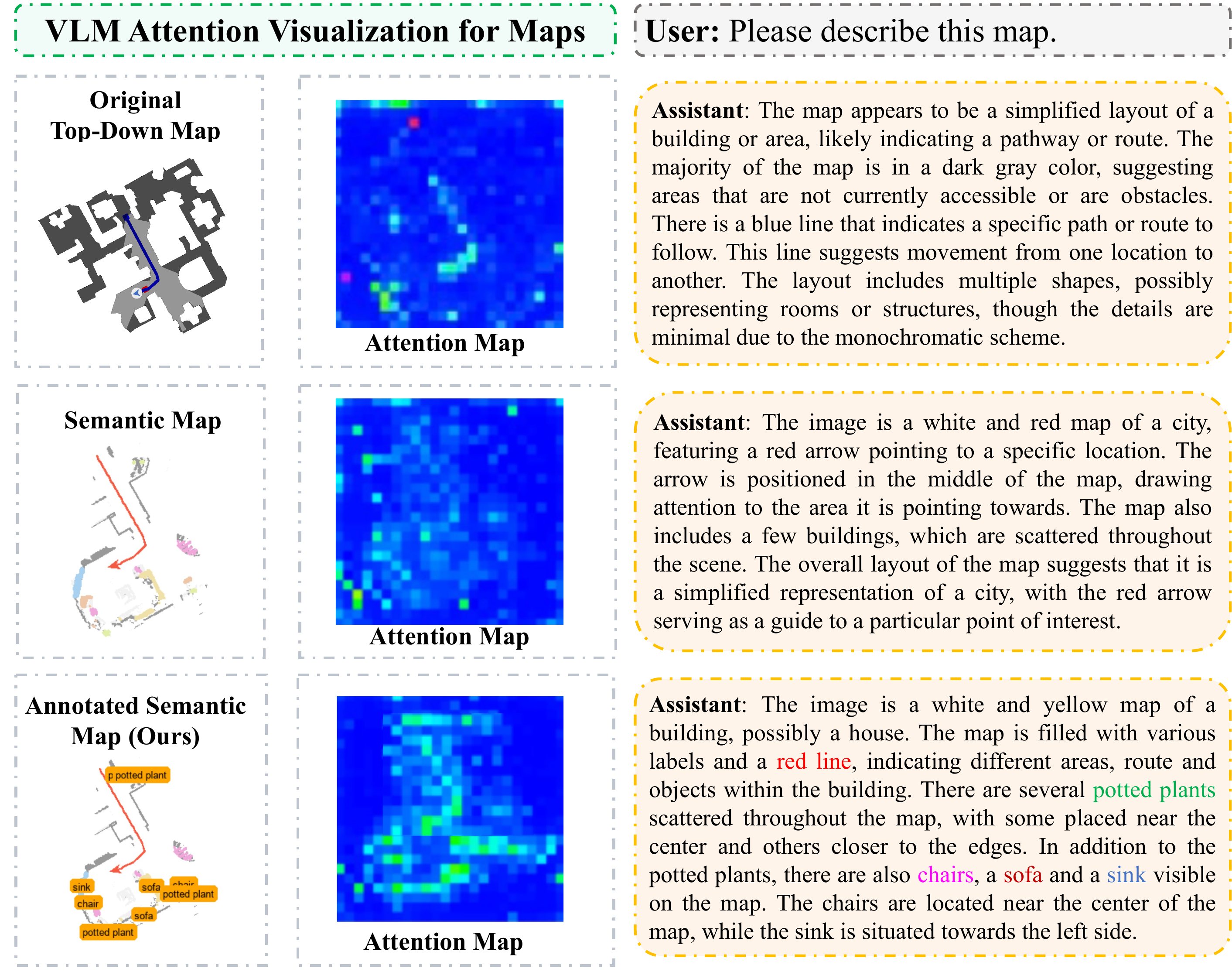}
\caption{\textbf{Additional Visualizations of VLM Attention Across Different Map Representations.}}
\label{figattnmap2}
\end{figure*}

\section{More Real-World Demos} 
\label{sec:append_realdemo}
To extensively validate our approach in real-world settings, we conducted comprehensive experiments and selected six representative episodes for visualization as shown in Fig.~\ref{fig:append_realdemo1} and Fig.~\ref{fig:append_realdemo2}. These episodes span five distinct real-world environments: office, meeting room, lecture hall, tea room, and living room, encompassing both simple and semantic instructions. The successful generation of ASMs in these diverse real-world scenarios and the agent's ability to complete assigned tasks demonstrate our system's effective sim-to-real transfer capabilities and the practical applicability of ASM in real-world navigation. Additionally, we conducted an experiment in a living room setting to test our VLM's semantic understanding capabilities. When presented with sofas of different colors, the agent was instructed to locate and navigate to the gray sofa. The successful execution of this task validates our model's advanced semantic understanding and its ability to differentiate between visually similar objects based on specific attributes.

\section{VLM Attention Visualization Analysis}
\label{attn}

To quantitatively evaluate our Annotated Semantic Map (ASM) representation, we conducted attention visualization analysis using VLM Visualizer\footnote{https://github.com/zjysteven/VLM-Visualizer} across different map representations. Using Vicuna-13B~\cite{zheng2023judging} as our base model, we generated attention heatmaps to examine how the model attends to different regions of the input maps. As shown in Fig.\ref{figattnmap} and Fig.\ref{figattnmap2}, the attention visualization reveals that when processing ASMs, the model exhibits significantly stronger attention alignment with semantically meaningful regions, showing sharp attention peaks ($>0.8$) precisely aligned with labeled objects and navigation-relevant areas. Moreover, our attention maps demonstrate that the red trajectory lines in ASM receive substantial attention focus, providing structured navigational cues that complement the semantic understanding. In contrast, attention patterns remain notably diffuse when processing the original top-down map (peak attention $<0.3$) and semantic map (peak attention $<0.4$), suggesting limited semantic understanding. This quantitative difference in attention patterns is further reflected in the model's descriptive outputs - while basic map descriptions indicate only geometric recognition (``a black and white map with a blue line'') for the original format, ASM enables sophisticated spatial-semantic understanding with precise object localization (\textit{e.g.}, ``bed in the upper left part''). These visualization results demonstrate that ASM's explicit textual annotations and structured trajectory representation successfully guide the model's attention to both semantic objects and navigation paths, enabling effective grounding of navigational features through the model's pre-trained language understanding capabilities.

\clearpage

\begin{figure*}[!ht]
\setlength{\abovecaptionskip}{-0.0001em}
\centering
\includegraphics[width=\textwidth]{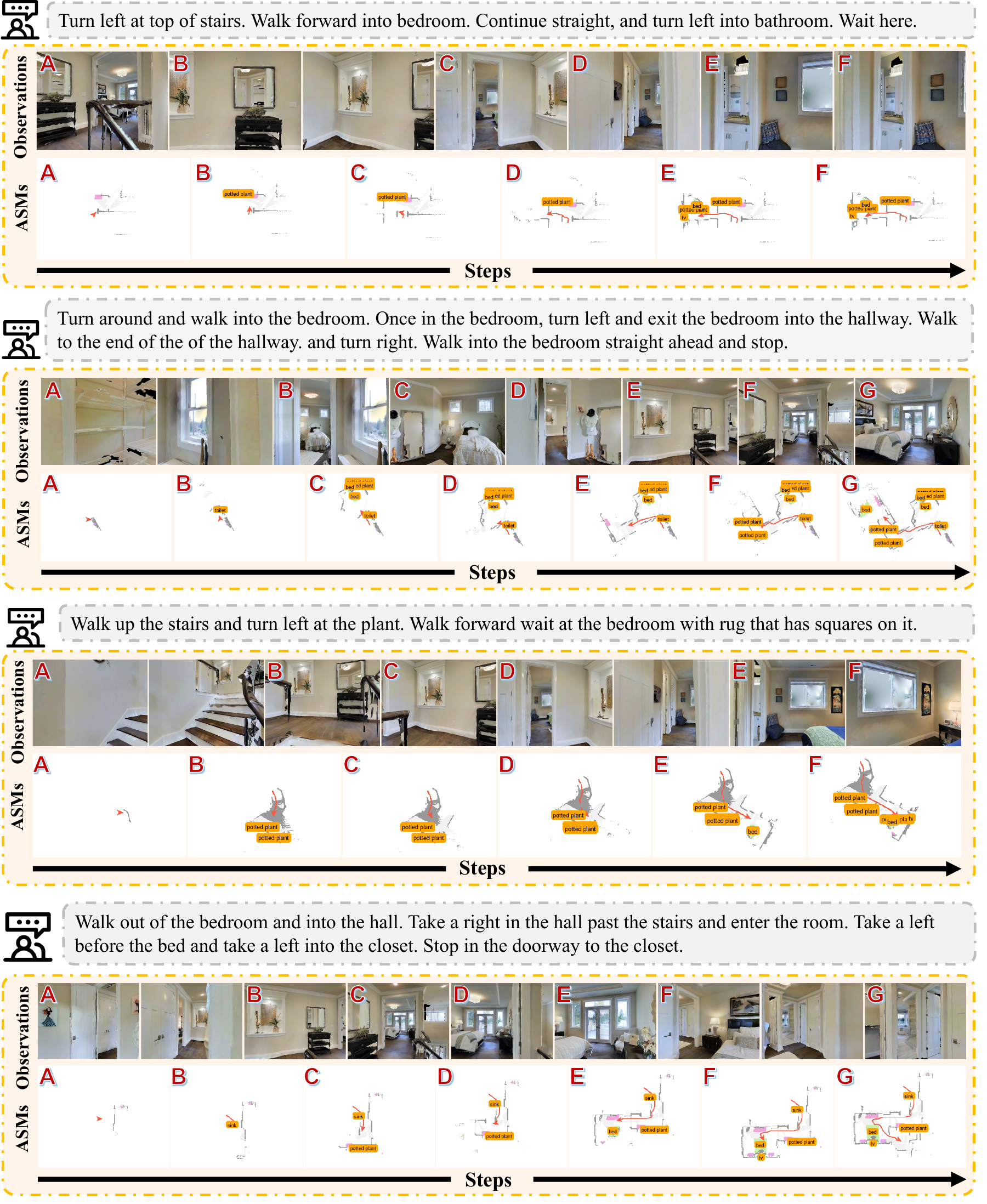}
\vspace{-10pt}
\caption{(1/6) Simulator demo results visualization.}
\label{fig:append_simdemo1}
\end{figure*}

\newpage
\begin{figure*}[t]
\setlength{\abovecaptionskip}{-0.0001em}
\centering
\includegraphics[width=0.99\textwidth]{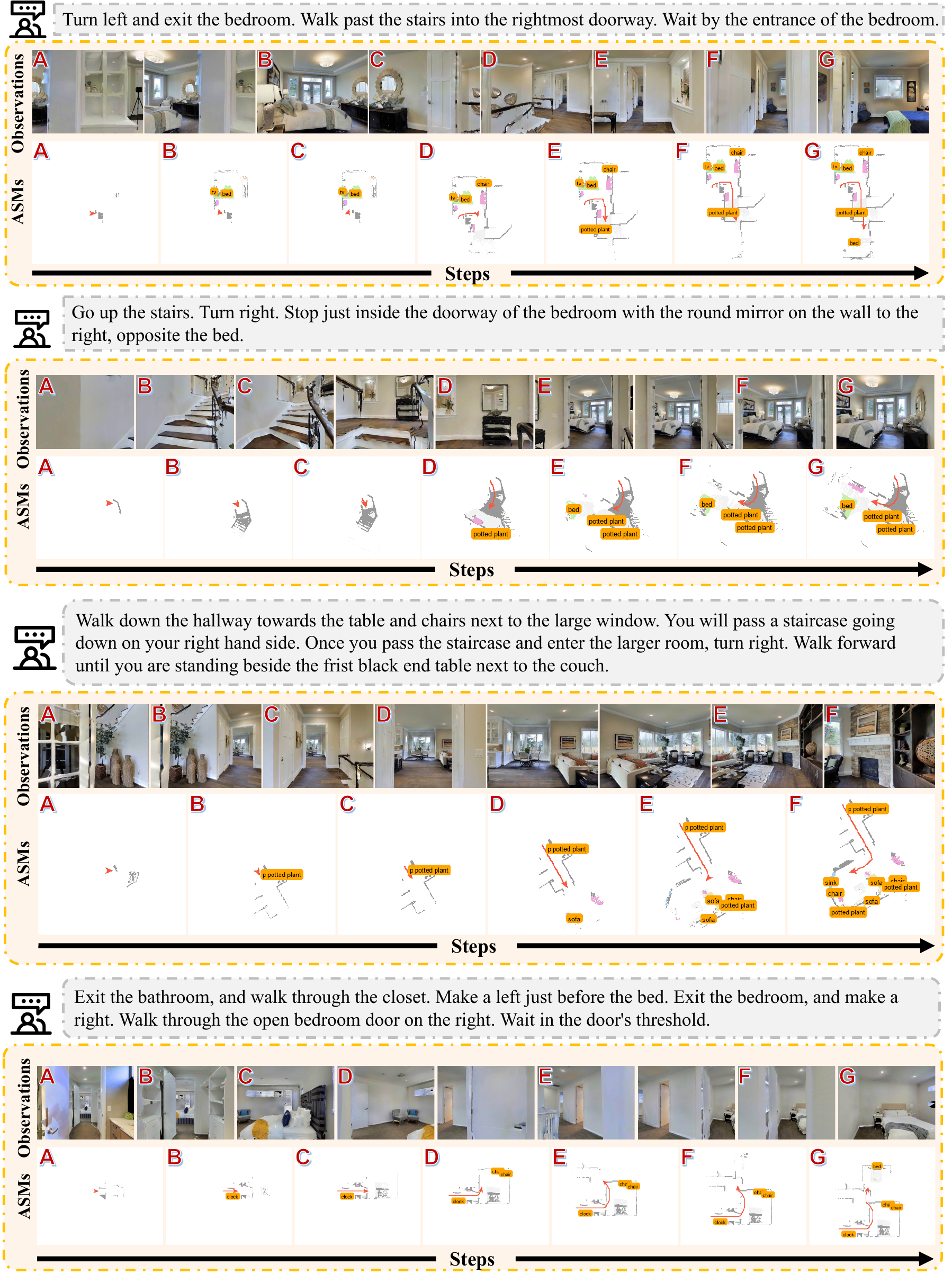}
\vspace{10pt}
\caption{(2/6) Simulator demo results visualization.}
\label{fig:append_simdemo2}
\end{figure*}

\newpage
\begin{figure*}[t]
\setlength{\abovecaptionskip}{-0.0001em}
\centering
\includegraphics[width=\textwidth]{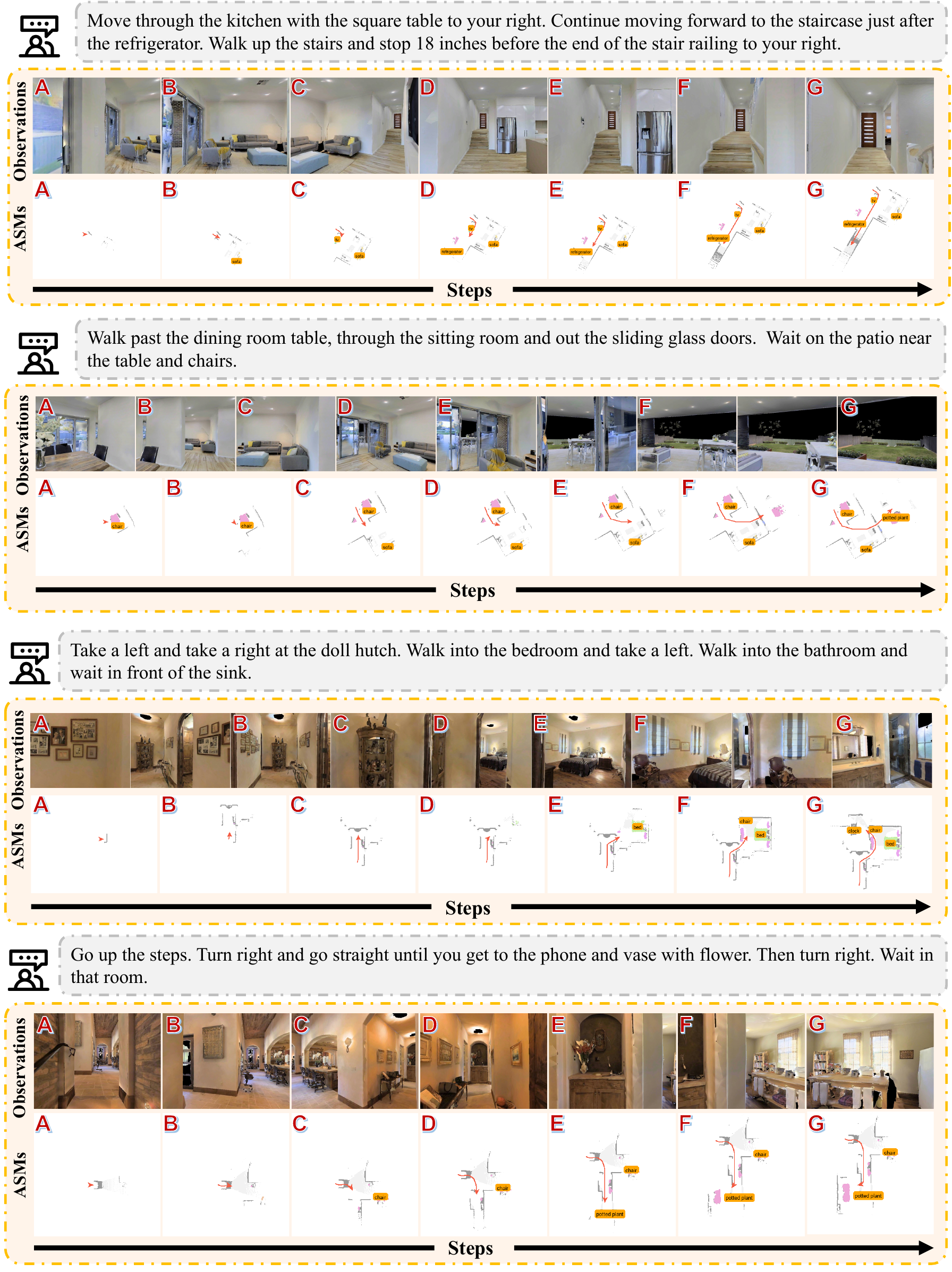}
\vspace{10pt}
\caption{(3/6) Simulator demo results visualization.}
\label{fig:append_simdemo3}
\end{figure*}

\newpage
\begin{figure*}[t]
\setlength{\abovecaptionskip}{-0.0001em}
\centering
\includegraphics[width=0.93\textwidth]{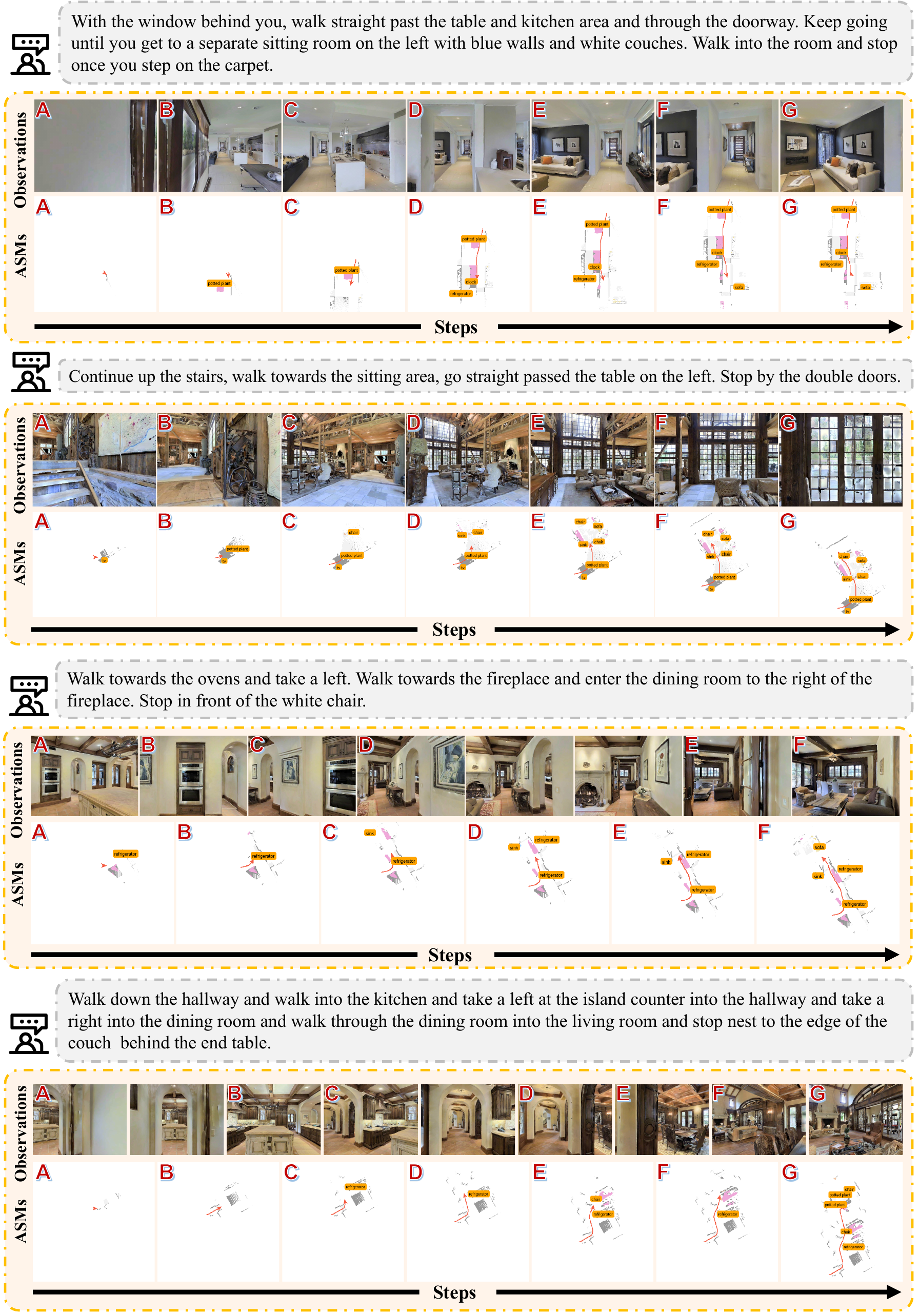}
\vspace{10pt}
\caption{(4/6) Simulator demo results visualization.}
\label{fig:append_simdemo4}
\end{figure*}

\newpage
\begin{figure*}[t]
\setlength{\abovecaptionskip}{-0.0001em}
\centering
\includegraphics[width=\textwidth]{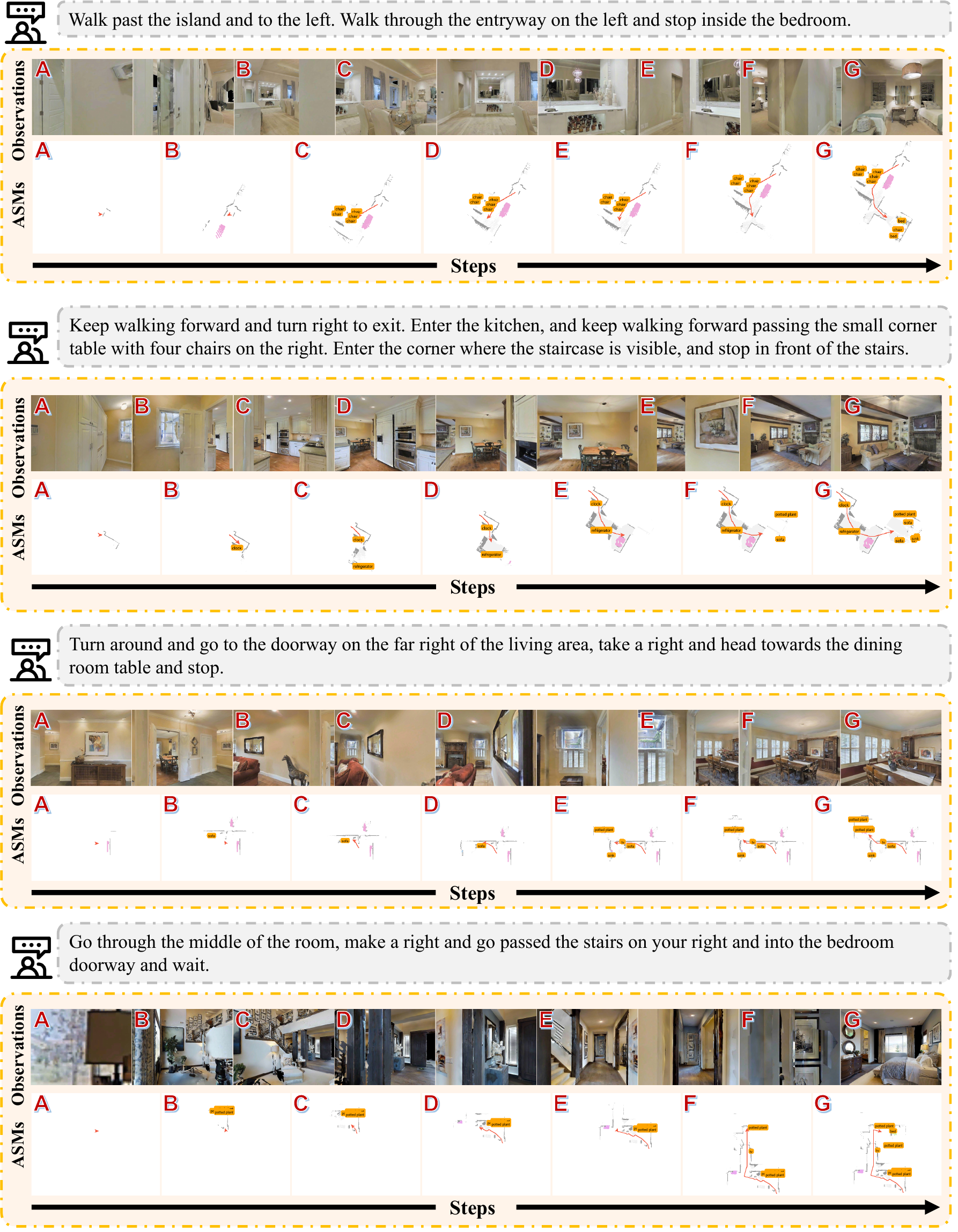}
\vspace{10pt}
\caption{(5/6) Simulator demo results visualization.}
\label{fig:append_simdemo5}
\end{figure*}

\newpage
\begin{figure*}[t]
\setlength{\abovecaptionskip}{-0.0001em}
\centering
\includegraphics[width=\textwidth]{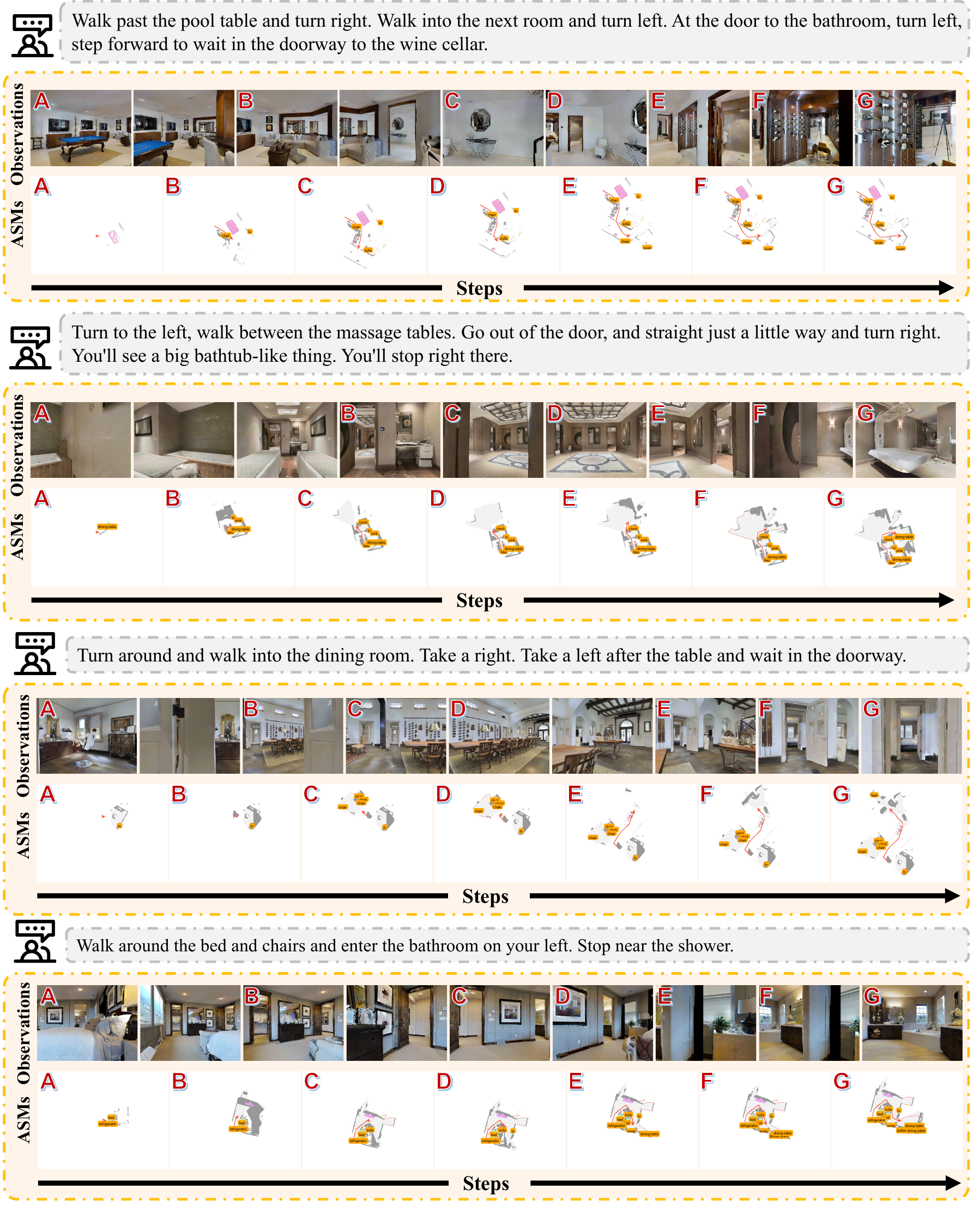}
\vspace{10pt}
\caption{(6/6) Simulator demo results visualization.}
\label{fig:append_simdemo6}
\end{figure*}

\clearpage


\begin{figure*}[!ht]
\setlength{\abovecaptionskip}{-0.0001em}
\centering
\includegraphics[width=\textwidth]{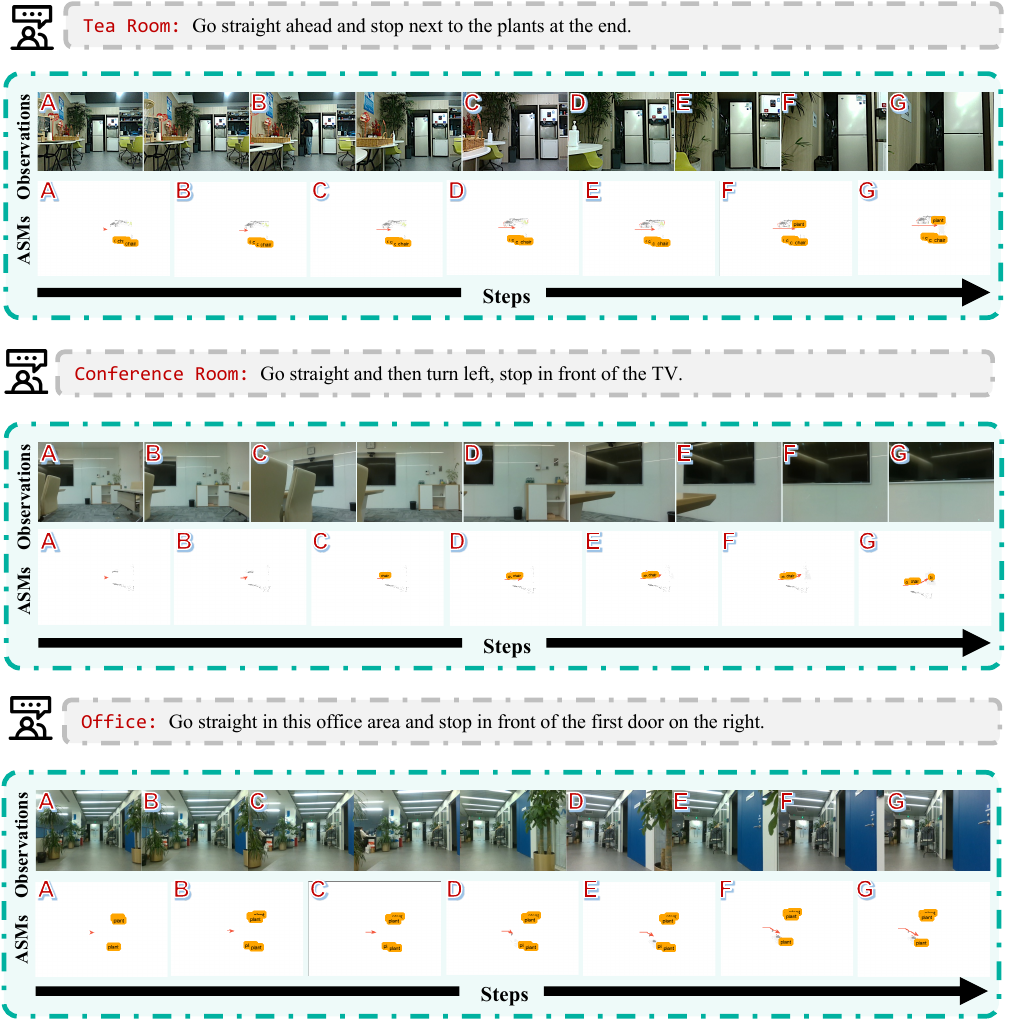}
\vspace{10pt}
\caption{(1/2) Real-world demo results visualization.}
\label{fig:append_realdemo1}
\end{figure*}

\newpage
\begin{figure*}[t]
\setlength{\abovecaptionskip}{-0.0001em}
\centering
\includegraphics[width=\textwidth]{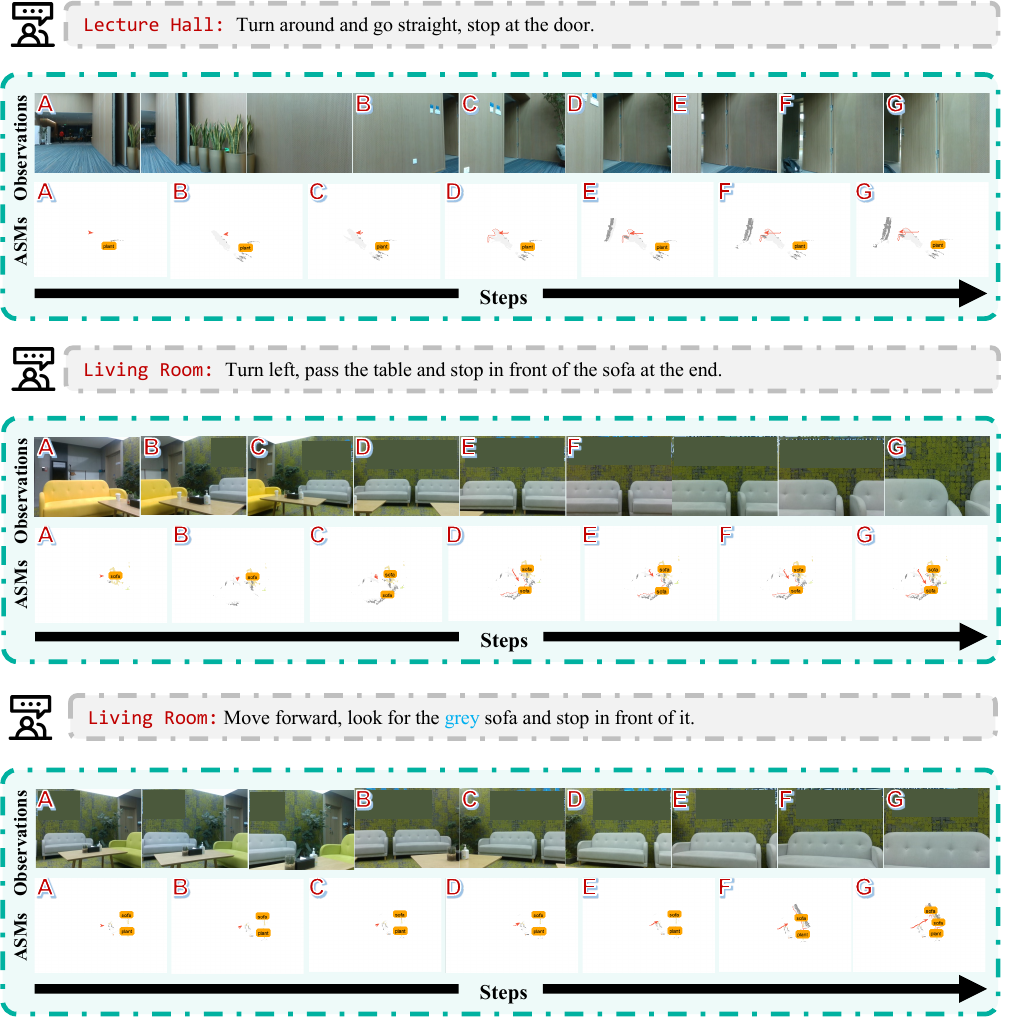}
\vspace{10pt}
\caption{(2/2) Real-world demo results visualization.}
\label{fig:append_realdemo2}
\end{figure*}

\end{document}